%% file: acl_latex.tex
\pdfoutput=1

\documentclass[11pt]{article}

\usepackage{acl}

\usepackage{times}
\usepackage{latexsym}

\usepackage[T1]{fontenc}

\usepackage[utf8]{inputenc}

\usepackage{microtype}

\usepackage{inconsolata}

\usepackage{graphicx}

\usepackage{algorithm}
\usepackage{algorithmic}
\usepackage{booktabs}
\usepackage{multirow}
\usepackage{multicol}
\usepackage{amsmath}
\usepackage{subcaption}
\usepackage{enumitem}
\usepackage{comment}

\newcommand{\premind}{{\fontfamily{lmtt}\selectfont PreMind}}

\title{\premind: Multi-Agent Video Understanding for Advanced Indexing of {\fontfamily{lmtt}\selectfont Pre}sentation-style Videos}

\author{
Kangda Wei\textsuperscript{\ddag}\thanks{\scriptsize{Work done during an internship at Bosch Research North America.}}, Zhengyu Zhou, Bingqing Wang, Jun Araki,\\ 
\textbf{Lukas Lange}, \textbf{Ruihong Huang\textsuperscript{\ddag}}, \textbf{Zhe Feng} \\ 
Bosch Research North America \& Bosch Center for Artificial Intelligence (BCAI)\\
\textsuperscript{\ddag}Department of Computer Science and Engineering, Texas A\&M University \\
\normalsize{\texttt{kangda@tamu.edu, \{Zhengyu.Zhou2, Bingqing.Wang, Jun.Araki\}@us.bosch.com,}} \\ 
\normalsize{\texttt{Lukas.Lange@de.bosch.com, huangrh@cse.tamu.edu, Zhe.Feng2@us.bosch.com}}
}

\begin{document}
\maketitle

\begin{abstract}
In recent years, online lecture videos have become an increasingly popular resource for acquiring new knowledge. Systems capable of effectively understanding/indexing lecture videos are thus highly desirable, enabling downstream tasks like question answering to help users efficiently locate specific information within videos. This work proposes \premind, a novel multi-agent multimodal framework that leverages various large models for advanced understanding/indexing of presentation-style videos. \premind \space first segments videos into slide-presentation segments using a Vision-Language Model (VLM) to enhance modern shot-detection techniques. Each segment is then analyzed to generate multimodal indexes through three key steps: (1) extracting slide visual content, (2) transcribing speech narratives, and (3) consolidating these visual and speech contents into an integrated understanding. Three innovative mechanisms are also proposed to improve performance: leveraging prior lecture knowledge to refine visual understanding, detecting/correcting speech transcription errors using a VLM, and utilizing a critic agent for dynamic iterative self-reflection in vision analysis. Compared to traditional video indexing methods, \premind \space captures rich, reliable multimodal information, allowing users to search for details like abbreviations shown only on slides.
Systematic evaluations on the public LPM dataset and an internal enterprise dataset are conducted to validate \premind's effectiveness, supported by detailed analyses.
\end{abstract}

\section{Introduction}

Recent technological advancements have led to the proliferation of online videos, which increasingly become an important source for learning new knowledge \cite{Soni}. Presentation-style lecture videos, which mainly present slides sequentially, are widely used for online courses and trainings \cite{10170589}. Systems that can effectively understand and index rich content of such videos thus become desirable \cite{10170589}, which could lead to advanced downstream applications such as question answering (QA) based on video details in various modalities. However, state-of-the-art (SOTA) approaches for indexing video content \cite{iyer2019contentbasedvideoindexingretrieval, unknown} remain unsatisfactory, as they fail to capture detailed multimodal information.  With the rapid advancements in large language model (LLM) technology \cite{zhao2023survey, chang2024survey}, Video Large Language Models (Vid-LLMs) \cite{lin2023videollavalearningunitedvisual, geminiteam2024geminifamilyhighlycapable} have emerged, enabling users to directly ask questions about provided videos \cite{pan2023retrieving}. However, Vid-LLM cannot answer questions that require systems to find the answers from a large number of videos due to its design and computation limitation.    

In this work, we propose \premind, a novel multi-agent multimodal framework that leverages various large models to capture detailed multimodal information in presentation-style lecture videos, leading to information-rich indexes that can benefit downstream tasks such as QA. For this work, we adopt the broad understanding of agents, viewing agents as system components that each has its own goals and work together to achieve a common goal \cite{Wang_2024}. \premind \space begins with a video segmentation component that combines a SOTA vision-based approach for segmentation with VLM to efficiently and reliably segment a video into many video segments, each covering one presentation slide. Then, \premind \space generates textual description for each segment with an advanced video-segment understanding component. For each segment, the component leverages appropriate agents to understand visual information, capture speech narrative, and generate a consolidated description. The component also involves innovative mechanisms to (i) improve vision understanding by leveraging knowledge learned previously in the video lecture, (ii) automatically correct speech recognition errors with VLM based on visual information and speech transcript, and (iii) further improve vision understanding through dynamic self-refinement with a critic agent. Based on the information-rich indexes created, downstream tasks such as retrieval-based QA and summarization can be implemented for various applications. 

We evaluate \premind \space using public LPM dataset \cite{Lee_2023_ICCV} as well as an enterprise internal dataset. Both intrinsic evaluation and extrinsic evaluation are conducted. The evaluation results show the effectiveness of \premind, demonstrating the value of capturing detailed multi-modal information in indexes, as well as the benefits of the proposed mechanisms
for the video understanding/indexing task.
To summarize, our contributions include:
\begin{itemize}[noitemsep, topsep=0pt, leftmargin=*]
    \item We introduce \premind, a multi-agent multimodal framework that uses various large models to capture accurate and detailed multi-model information from presentation-style lecture videos, which can further benefit the possible downstream applications such as QA.
    \item We demonstrate the effectiveness of \premind \space on the public LPM dataset \cite{Lee_2023_ICCV} and an enterprise internal dataset through intrinsic and extrinsic evaluations. 
    \item We conduct ablation studies to evaluate the efficacy of different mechanisms within \premind, and present comprehensive analyses of framework's efficiency as well as case studies.
\end{itemize}

\section{Related Work}
\paragraph{Lecture Video Indexing}
Indexing lecture videos is an increasingly crucial task for enhancing access to relevant content within educational materials, which often involves video segmentation and information extraction from video segments. For segmentation, \citet{9378632} used Voice Activity Detection and Gaussian Mixture Models to segment videos based on speech. \citet{Shah2015TRACELA} aligns spoken content with Wikipedia text for better segmentation. \citet{Jeong2012AnAL} applied SIFT for precise slide detection.
For information extraction, Optical Character Recognition (OCR) helps retrieve text from slides and Automatic Speech Recognition (ASR) helps obtaining the speech in text format. The extracted text is then used for indexing the video in a content-based manner. \citet{10.1145/267437.267478} used OCR for hierarchical indexing, while \citet{6750040} combined ASR and OCR for comprehensive video search. Multimodal approaches, like those by \citet{DBLP:conf/interspeech/YamamotoOA03} and \citet{1265045}, integrate ASR and OCR for improved indexing. However, indexes generated from previous works \cite{yang2011lecture, 8260637, Yang2011LectureVI, Debnath2023AML, Arazzi2023SemanticHI, medida2021video} do not have rich multi-modal information but rather simple text description obtained from OCR or ASR.
Despite advancements, challenges remain in achieving high accuracy in indexing due to ASR errors, visual variations, and lecture complexity. However, these issues can be mitigated with the help of LLMs and VLMs. To the best of our knowledge, our proposed framework is the first to utilize Large Models to enhance video indexing quality.

\paragraph{Video Large Language Models}
Vid-LLMs are widely used for tasks involve video understanding \cite{abdullah2024ualbenchcomprehensiveunusualactivity}.
For Vid-LLM, it typically uniformly samples a certain numbers of frames from videos and utilize a visual encoder \cite{Dosovitskiy2020AnII, radford2021learningtransferablevisualmodels} to convert each frame into vector representation. Then, an adapter is used to map the video embeddings from visual semantic space to text semantic space of LLMs. Textual embeddings of instructions are then added generating responses for downstream tasks \cite{zhang-etal-2023-video, maaz-etal-2024-video, lin-etal-2024-video, song2024moviechatdensetokensparse, chen2023videollmmodelingvideosequence, Ma_2024_CVPR}, or a specially designed task-specific head can be used to perform regression tasks \cite{yu2023selfchainedimagelanguagemodelvideo, Huang_2024_CVPR, Ren_2024_CVPR, li-etal-2024-groundinggpt}. For Vid-LLMs, although previous works have demonstrated impressive video understanding capabilities, they are not well-suited for content-based lecture video indexing. One limitation lies in their sampling strategy, which is suboptimal as it can result in redundant information or the omission of critical details \cite{wang2024videotreeadaptivetreebasedvideo}, thus not suitable for video indexing that requires segmentation. Additionally, these systems are mostly designed to process one query about one video or several at a time in an online fashion, thus not suitable for indexing videos.

\begin{figure*}[h]
    \centering
    \includegraphics[scale=0.48, trim={1cm 1.5cm 1cm 2.5cm}]{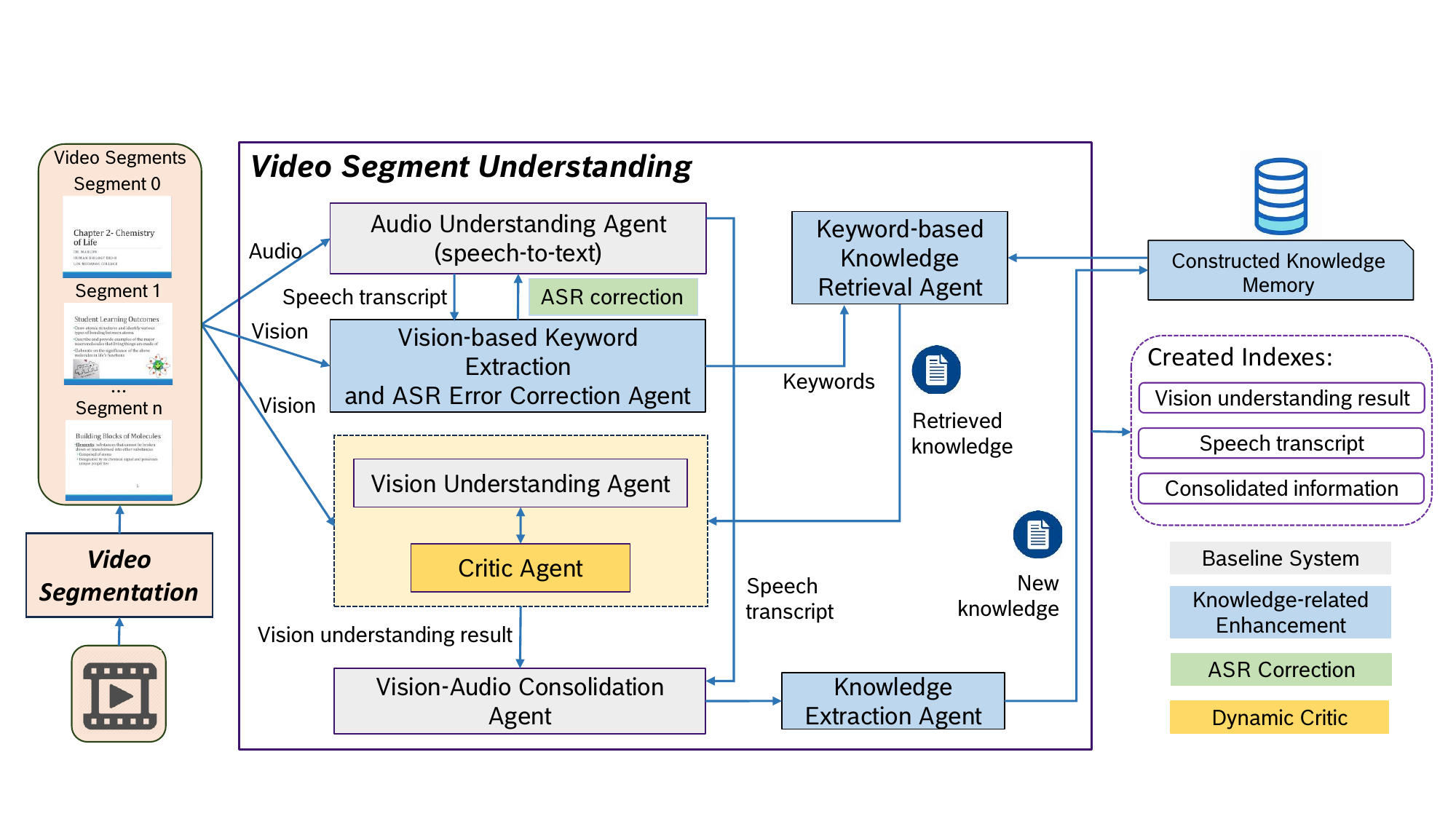}
    \vspace{-0.2cm}
    \caption{Illustration of the proposed \premind \space framework. 
    }
    \label{fig:main}
    \vspace{-0.4cm}
\end{figure*}

\section{Method}
The overall structure of \premind \space is illustrated in Figure~\ref{fig:main}.  It consists of a video segmentation component and a video-segment understanding component, generating three understanding results based on vision, speech, and consolidated information, respectively, as the output indexes.    
Given a number of lecture videos, \premind \space processes the videos one by one, and the resulting pool of indexes can be used in downstream tasks.

\subsection{Video Segmentation}

The video segmentation component attempts to segment a training video into multiple segments, each covering the presentation of one slide.  
We adopt the state-of-the-art PySceneDetect \cite{gruzman2014algorithm,reddy2015comparison} to conduct the first-round segmentation. PySceneDetect faces two major challenges for this task, (1) often missing slide with similar layout as the precedent one and (2) splitting the presentation of a single slide into multiple segments due to background changes. Therefore, we innovatively use VLM to refine the segmentation when needed. After the first-round segmentation, for long segments (>1 minute) detected by PySceneDetect, we apply VLM to re-detect slides ($Step_A$) in that segment with the aim of catching the missed slides.  The time span for each newly detected slide is determined using vision and audio cues ($Step_B$).  For other segments, VLM is leveraged to merge the current segment with the previous one if the two segments are deemed as presenting a same slide. More details of our proposed video segmentation algorithm is shown in Appendix~\ref{sec:appendix_video_segmetation}.

\subsection{Video-Segment Understanding}

With the obtained video segments, denoted as $\{S_1, S_2, ..., S_n\}$, the video-segment understanding component attempts to understand/index the content of each segment $S_i$ with a  multi-agent solution, as shown in Figure 1.  We develop the understanding component in an incremental way.  First, we design a baseline understanding system, which adopts three separate agents to extract audio, vision, and consolidated information from the segment. We then leverage knowledge to enhance the vision understanding, extracting knowledge per slide, keeping a knowledge memory for the lecture in focus, and leveraging the previously-learned knowledge to understand the current slide. We further reduce the impact of ASR
errors on understanding results by leveraging VLM to automatically correct ASR errors based on both speech transcription and slide visual content.  Finally, we introduce a critic agent to dynamically refine vision understanding result in a self-reflection manner. 
Prompt templates for all agents/algorithms can be found in Appendix~\ref{sec:appendix_prompts}.

\subsubsection{Baseline System}
\begin{itemize}[noitemsep, topsep=0pt, leftmargin=*]
  \item \textbf{Audio Understanding Agent:} This agent uses a ASR model to generate a speech transcript, denoted as $transcript_i$, for each segment $S_i$. In practice, the ASR model is applied to transcribe the whole video in focus. After the video segmentation process, the audio understanding agent extracts $transcript_i$ for $S_i$ from the whole transcript based on the starting/ending times of $S_i$. 
  \item \textbf{Vision Understanding Agent:} Given $S_i$, this agent generates a detailed description of the slide presented in the segment using a SOTA VLM.  It samples a (representative) video frame from $S_i$ and asks the VLM to describe the slide shown in that frame image in detail. The description generated is the vision understanding result, denoted as $vision\_understanding_i$.
  \item \textbf{Vision-Audio Consolidation Agent:} Based on the audio and vision understanding results, the consolidation agent further generates a consolidated understanding result, denoted as $consolidated\_information_i$, to provide a good overall understanding about what is presented in the video segment $S_i$.   
\end{itemize}

\subsubsection{Knowledge-Related Enhancement}
\begin{itemize}[noitemsep, topsep=0pt, leftmargin=*]
  \item{\textbf{Knowledge Memory}}: Given a segment $S_i$, knowledge presented in previous slides may be helpful to understand the current slide. 
  We maintain a knowledge memory $KM$ for the lecture that $S_i$ is part of. The knowledge memory contains entries in the following format: $knowledge_m$ = \{$embedding_m$, $name_m$, $explanation_m$\}, where $m \in M$ and $M$ is the number of entries in $KM$. In each entry, $name_m$ stores a concept name, such as Product Lifecycle Management, $explanation_m$ stores the explanation of that concept, and $embedding_m$ stores the embedding representation of the concept name, which is obtained by a SentenceTransformer model\footnote{\url{https://www.sbert.net/}}, for knowledge retrieval. For each lecture, $KM$ is initially empty. It is then updated when new knowledge is extracted, starting from the first segment of the lecture. Notice that, unlike previous works described in \cite{Hatalis_Christou_Myers_Jones_Lambert_Amos-Binks_Dannenhauer_Dannenhauer_2024}, where memory is used for tasks that only involve text, we keep a knowledge memory that contain multi-modal information from previous video segments to help understand current segment.
  \item \textbf{Knowledge Extraction Agent:} This agent extracts new knowledge from the consolidated understanding result for $S_i$, and updates $KM$ correspondingly. It asks a SOTA LLM to extract concepts, each including a concept name and an explanation, from $consolidated\_information_i$. For each extracted concept, embedding representation of the concept name, $e$, is computed with the SentenceTransformer, and entries in $KM$ are ranked by cosine similarity between $e$ and $embedding_m$. If the top-ranked entry has a similarity score less than 0.7, the extracted concept is deemed new and we update the $KM$ by inserting the concept as a new entry.  Otherwise, the top-ranked entry is deemed to have similar concept with the extracted concept. In this case, we update the top-ranked entry $knowledge_j$ by (1) appending the explanation of the extracted concept to $explanation_j$\ and (2) updating $embedding_j$ as the rolling average of $embedding_j$ and $e$.
  
  \item \textbf{Keyword Extraction (Part of the Keyword Extraction and ASR Error Correction Agent):} Given  $S_i$, we extract  a set of keywords, denoted as $keywords_i$, from the slide vision using a VLM. These keywords are then used to retrieve relevant knowledge from $KM$. 
  In this work, we merge the tasks of keyword extraction and ASR error correction into one agent, using one VLM prompt to accomplish the two tasks at the same time for improved efficiency and performance.  
  
  \item \textbf{Keyword-based Knowledge Retrieval Agent:} With $keywords_i$ extracted from $S_i$, this agent  retrieves relevant knowledge from $KM$ to facilitate vision understanding. For each keyword in $keywords_i$, the agent computes its embedding representation using the SentenceTransformer model, and then ranks the $KM$ entries by cosine similarity between the keyword's embedding and $embedding_m$.  Among the top 10 entries, those entries with  similarity score larger than 0.7 are deemed relevant, and are provided to the vision understanding agent as context.  
\end{itemize}

\subsubsection{ASR Correction}
Mistakes made by ASR model have a negative impact on the quality of generated understanding results.
To reduce ASR errors, this work proposes an innovative approach that leverages reliable visual information, i.e., the keywords extracted from slide vision, to correct ASR errors using a VLM, while previous works on ASR correction mainly rely on ASR results alone \cite{ ma2023generativelargelanguagemodels, ma2025asrerrorcorrectionusing}.
For the keyword extraction and ASR error correction agent, given $S_i$, we use one prompt to ask the VLM to (1) extract keywords shown in the slide, and (2) check $transcript_i$ to identify possible ASR mistakes made on the keywords and make correction suggestions.
The combination of the two tasks not only enhances efficiency but also benefit the performance of ASR correction, as constraining the correction scope to keywords helps reduce hallucination of VLM on ASR correction. 

\subsubsection{Dynamic Critic}
LLM self-reflection has been found effective for various text processing tasks\cite{madaan2023selfrefineiterativerefinementselffeedback, liang-etal-2024-encouraging}.  In this work, we extend this technique to multi-modal data, introducing a critic agent to enhance vision understanding through iterative reflection.
Given the slide of $S_i$, the retrieved knowledge, and the $vision\_understanding_i$ generated by the Vision Understanding agent, the critic agent aims to identify defects of $vision\_understanding_i$, such as counting mistakes and missing figures in description. With the feedback from the critic agent, the vision understanding agent further improves the understanding result and sends the updated result to the critic agent to review. This process iterates until the critic agent is satisfied with the understanding result. We realize this dynamic reflection mechanism by grouping the critic agent and the vision understanding agent into a AutoGen \cite{wu2023autogenenablingnextgenllm} groupchat, which also includes an admin agent to assist the chatting function.  We configure the groupchat to allow at maximum $N_{max}$ vision-understanding/critic calls in the reflection iteration. For early termination, we define the chat termination condition as the 'TERMINATE!!!' command issued by the Critic Agent. 

\section{Experiments and Analyses}
\subsection{Dataset}
\vspace{-0.2cm}
\input{latex/data_stats_segmentation}
\vspace{-0.2cm}
\input{latex/data_stats_understanding}
We evaluate \premind \space on the public LPM dataset \cite{Lee_2023_ICCV} and an enterprise internal (EI) dataset. The LPM dataset contains YouTube lectures across 10 different categories (e.g., biology, psychology), having more than 180 hours of videos and providing manually- segmented slide-presentation segments for over 9,000 slides in total. The internal dataset contains 66 videos (6 hours in total) on various enterprise-training topics. The videos in both datasets are presentation-style lectures, though the layout of the slide display in the videos varies. 
In this work, we randomly sample a subset of videos per dataset to evaluate video segmentation performance, as shown in Table~\ref{tab:data-stats-video-segmentaion-table}. Note that for the LPM videos sampled (listed in Appendix~\ref{sec:appendix_lpm_videos}), the first 10 minutes of each video is used in video-segmentation evaluation. For the evaluation of understanding performance, we select another subset of lectures from the LPM data, which contains almost 30 hours of videos in total, due to computational constraints. We use this LPM subset and the whole EI dataset to evaluate the video-understanding approaches as well as QA performance in extrinsic evaluation, as listed in Table~\ref{tab:data-stats-video-segment-understanding-table}. We construct the LPM subset for understanding evaluation by (1) selecting the first three lecture videos from each category except dental, and (2) for dental videos, which contain 19 subcategories and each video is only around 5 minutes, selecting all the videos of the first 3 subcategories.

\subsection{Video Segmentation Evaluation}
\subsubsection{Settings}
We evaluate our proposed video segmentation approach and compare it with PySceneDetect in Table~\ref{tab:data-stats-video-segmentaion-table}. Details on the parameter tuning as well as algorithm configurations are provided in Appendix~\ref{subsec:appendix_system_settings_videp_segmentation}. GPT-4 Vision is used in our proposed segmentation algorithm. For video-segmentation evaluation, we report Precision, Recall, and F-1 score for detecting video segments. We also report the Intersection over Union (IoU) score (ranging from 0 to 1), which specifies the amount of overlap on time span between the predicted and ground truth segments. IoU is calculated as:

\vspace{-0.1cm}
\begin{equation}
    \scalebox{0.7}{$
        \begin{aligned}
            IoU & = \frac{|A \cap B|}{|A \cup B|}
        \end{aligned}
    $}
    \label{eq1}
\end{equation}
\vspace{-0.1cm}

where $A$ and $B$ are the time spans of the the predicted and ground-truth segments, respectively.

\subsubsection{Video Segmentation Results}
\input{latex/segmentation_table}
The evaluation results on video segmentation are shown in Table~\ref{tab:segmentaion-table}. We can see that our proposed segmentation approach significantly outperforms SOTA vision-based approach, demonstrating the power of VLM.  Our approach achieves almost perfect results on EI, successfully detecting all presented slides. The performance on LPM is somehow imperfect, as the LPM data contains occasional occurrences of animations/demonstrations, making it more challenging.  From efficiency aspect, as our proposed segmentation approach only applies VLM when needed, the computational overhead introduced with VLM is minimized, and the segmentation efficiency is largely maintained (as will be further discussed in Section~\ref{sec:framework efficieny}). 

\subsection{Video-Segment Understanding Evaluation}
\subsubsection{Settings}
We conduct video-segment understanding experiments based on the datasets listed in Table~\ref{tab:data-stats-video-segment-understanding-table}. For these experiments, we directly use the provided manual segmentation for the LPM data and use our proposed segmentation algorithm to segment the EI data. Using video segments obtained in this manner, four incrementally developed video-segment understanding approaches are applied to each dataset: (1) the baseline system, (2) plus knowledge-related enhancement, (3) plus ASR correction, and (4) plus dynamic critic. Each approach is applied individually, generating a corresponding set of understanding results, which are then evaluated and compared to assess their performance. In this set of experiments, Whisper\footnote{\url{https://openai.com/index/whisper/}} is used to generate ASR result, and GPT-4 Turbo is used for all agents that require a VLM/LLM in processing. Algorithm parameters for video-segment understanding are determined empirically, with details provided in Appendix~\ref{subsec:appendix_system_settings_videp_understanding}.  

\subsubsection{Intrinsic Evaluation}

\paragraph{Evaluation Approach}
We first evaluate the four proposed video-segment understanding approaches on vision understanding performance.  As this is a challenging task by nature, we adopt human annotation to ensure the evaluation quality, and propose a pairwise comparison schema for evaluation. Given a pair of vision understanding results to compare together with the corresponding slide image, to reduce the labeling workload for human, we first ask GPT-4 Turbo to determine whether the two understanding results are consistent in meaning ( prompt listed in Appendix Table~\ref{tab:prompt-table3}). 

When an inconsistency is detected, the result pair is sent to humans for manual quality comparison. For each pair requiring manual evaluation, a questionnaire is prepared, asking annotators to determine which result is better and explain their reasoning based on the slide image and relevant knowledge. Each questionnaire is completed by three annotators, and the final judgment is determined by a majority vote. For EI data, due to confidential issue, we recruit internal associates to annotate related questionnaires. For the LPM data, we leverage Amazon Mechanical Turk \footnote{\url{https://www.mturk.com/}} for the annotation, designing special measures to ensure high quality of annotation (e.g., selecting workers having trustful records, using dummy questions to reject irresponsible answers).  The details about the measures and the questionnaire examples are provided in Appendix~\ref{sec:appendix_human_annotation}. 

\input{latex/human_eval_table}
\input{latex/main_table}

\paragraph{Results} 
Table~\ref{tab:human-eval-table} reports the comparison results on vision understanding performance for different video-segment understanding settings against the baseline system. For each pair of approaches in comparison, we list (1) the total number of cases where the final vision-understanding results generated by the two approaches are deemed different by GPT-4 Turbo, and (2) among those cases, which are sent to human annotators to label, the competition results according to the human annotation.  If the first approach is deemed better than the second approach according to annotation results, it wins the second approach.  If it is deemed similar to the second approach, a tie occurs. Otherwise, the first approach loses the competition.  From Table~\ref{tab:human-eval-table}, we can see that adding knowledge-related enhancement, ASR correction, and dynamic critic gradually improves the quality of vision understanding for both LPM and EI data.  Note that the improvement brought by ASR correction on vision understanding is indirect (i.e., better speech transcript $\rightarrow$ better consolidated info $\rightarrow$ better extracted knowledge $\rightarrow$ better knowledge-assisted vision  understanding). 
For lose cases, we suspect these are caused by the noises in the knowledge that is retrieved and fed into the vision understanding agent.  These noises may distract the agent from important information during the generation of slide description.

\subsubsection{Extrinsic Evaluation with QA}
\paragraph{Evaluation Approach}
We conduct extrinsic evaluation by applying the generated indexes for QA tasks to show the impact of the proposed video-segment understanding component on downstream application.  With the three indexes (i.e., vision understanding result, speech transcript, and consolidation information) generated for each video segment in a dataset, we setup a retrieval-augmented QA system.  
In this set of experiments, we adopt the complete video-segment understanding component (i.e., the B+K+A+D setting) to generate the understanding results as textual indexes. 
For evaluation, we ask GPT-4 Turbo to generate 6 questions together with the ground-truth answers. Among the 6 questions, 2 focus on speech, 2 focus on slide vision, and 2 focus on the consolidated information.  After using QA to generate an answer for each question, we further leverage GPT-4 Turbo to evaluate the answer based on the ground-truth. The details about the QA settings, question/ground-truth generation, and the GPT based evaluation are all provided in Appendix~\ref{subsec:appendix_system_settings_qa}.

\paragraph{Results}
We evaluate the QA performance as the accuracy of the answers generated by QA.  The results are reported in Table~\ref{tab:main-table}.  The performance of only using vision-understanding-result indexes, only using speech-transcript indexes, only using consolidated-information indexes, and using all types of indexes in retrieval are evaluated per question type.  The results demonstrate the value of capturing multimodal information in indexes for QA.  Each type of indexes (e.g., vision understanding result) is advantageous for certain kind of questions (e.g., those asking for vision details not presented by speech).  And adopting indexes of all types bring the best overall performance.

\subsubsection{Ablation Study}
\input{latex/speech_table_updated}
Leveraging the QA system, we further conduct ablation study on the benefit of the proposed ASR correction procedure by comparing the system performance with and without this procedure. For this evaluation, we generate another set of questions together with their ground-truth in a similar way as before.  In this case, based on the predicted ASR corrections, we ask GPT-4 Turbo to generate questions answerable with the corrected speech transcript but not answerable without the corrections (See Table~\ref{tab:prompt-table2} for the prompt). With this set of questions, using only speech-transcript indexes in the QA system, we evaluate the answer accuracy as before.  We focus on the speech-transcript indexes here for the easiness of interpretation, as the major impact of ASR Correction is on audio understanding result. The evaluation results are shown in Table~\ref{tab:speech-table}. We can see that ASR correction brings substantial improvements on QA accuracy for both datasets, demonstrating the benefit of correcting ASR errors on video understanding.    

\input{latex/vision_table_updated}
In a similar manner, we also conduct another ablation study on vision-understanding aspect. In this case, we focus on each pair of vision-understanding results that has one result determined as better than the other in human annotation, and ask GPT-4 Turbo to generate questions answerable with the better result but not with the other one (See Table~\ref{tab:prompt-table2} for the prompt). With these questions, we evaluate the four proposed video-segment understanding settings with QA, using only the vision-understanding indexes in the QA system in this case.  The results are reported in Table~\ref{tab:vision-table}.  We can see that both the knowledge-related enhancement and the dynamic critic mechanism bring substantial improvements on performance.  For ASR correction, which only has indirect impact on vision understanding, it unsurprisingly brings only mild improvement on EI, and brings no improvement on LPM probably due to generation noises.     

\subsubsection{Case Study}
\input{latex/case_study}
We also examine improvements achieved through knowledge enhancement, ASR correction, and dynamic critic separately. Table~\ref{tab:case-study-table} shows one example per dataset for each mechanism.  We observe that knowledge is useful to help VLM understand ambiguous part of slides. For example, on a slide, several stages of PLM (product lifecycle management) are listed right below the ERP (enterprise resource planning) block. Although the PLM stages are not connected to the ERP block in the slide, the VLM wrongly describes that ERP includes those stages. This problem is fixed when definitions of PLM and ERP are retrieved and provided to the VLM for vision understanding. The ASR correction procedure excels at refining domain-specific terms and resolving uncommon or ambiguous names. The dynamic critic self-reflection mechanism effectively addresses specific errors made by the VLM in slide descriptions, such as misinterpreting numbers in figures, making counting errors, and missing visual elements in the slide descriptions.

\subsection{Framework Efficiency}
\label{sec:framework efficieny}
\input{latex/efficiency_table}
We evaluate the efficiency of our proposed framework \premind \space in Table~\ref{tab:efficiency-table}. 
Note that the framework efficiency is largely determined by the API calls used in various components. 
For each video, one ASR API call is needed to generate a transcript for the whole video.  For the video segmentation procedure, except the VLM API calls, the additional processing time beyond the first-round PySceneDetect is negligible. For the video-segment understanding, API calls occupy the majority of processing time, while the average time for retrieving knowledge from the knowledge memory and the average time for correcting detected ASR errors per video segment are $364$ ms and $10$ ms, respectively. 

\section{Conclusion}
This work proposes \premind, a novel framework to understand/index rich multimodal information for presentation-style lecture videos, with the aim of enabling advanced downstream applications such as QA. 
\premind \space involves two components: video segmentation and video-segment understanding. The video segmentation procedure combines VLM with PySceneDetect to achieve the desired performance with high efficiency. The video-segment understanding component not only captures visual/audio/consolidated information from video using large models, but also introduces three innovative mechanisms
to improve the understanding performance. We evaluate \premind \space on the public LPM dataset and an internal dataset and encouraging experimental results are achieved.

\section*{Limitations}
\premind \space relies heavily on proprietary LLMs and VLMs for both video segmentation and understanding tasks. Open-sourced models may be used to substitute proprietary models, but the performance may be effected. \premind  \space is optimized for presentation-style lecture videos. Its generalizability to other video formats, such as freestyle videos without any slides presented, has not been explored. The reliance on human annotation for evaluation introduces subjectivity, particularly in vision understanding tasks where judgment about description quality can vary across annotators.
Future work will address these limitations by exploring lightweight and open-sourced model alternatives and expanding the evaluation to include more diverse datasets and video formats.
\bibliography{anthology, custom}

\newpage

\appendix
\section{Appendix}
\label{sec:appendix}
The Appendix consists of supplemental materials in our research journey for the video understanding topic. 

\subsection{Video Segmentation Details}
\label{sec:appendix_video_segmetation}
\begin{figure}[t]
    \centering
    \includegraphics[scale=0.65, trim={1cm 0cm 1cm 1cm}]{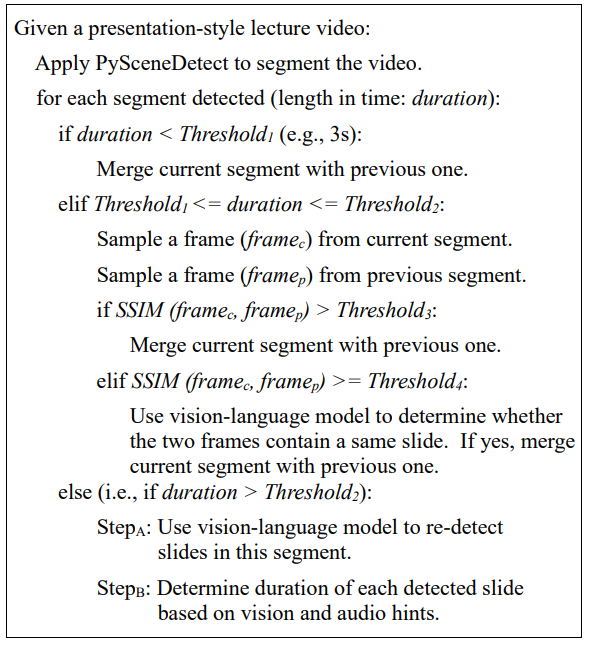}
    \caption{Video segmentation algorithm. SSIM is the Structure Similarity Index, a measure in image assessment, which gives values between 0 and 1. 
    SSIM gives high similarity, when two images have visually similar look but have rather different pixel value (e.g. stretch, mean-shift). While the approach is prone to give a low measure value, if visual appearance of the two images is much different. }
    \label{fig:video_segmentation_algorithm}
\end{figure}

\begin{figure*}[h]
    \centering
    \includegraphics[scale=0.9]{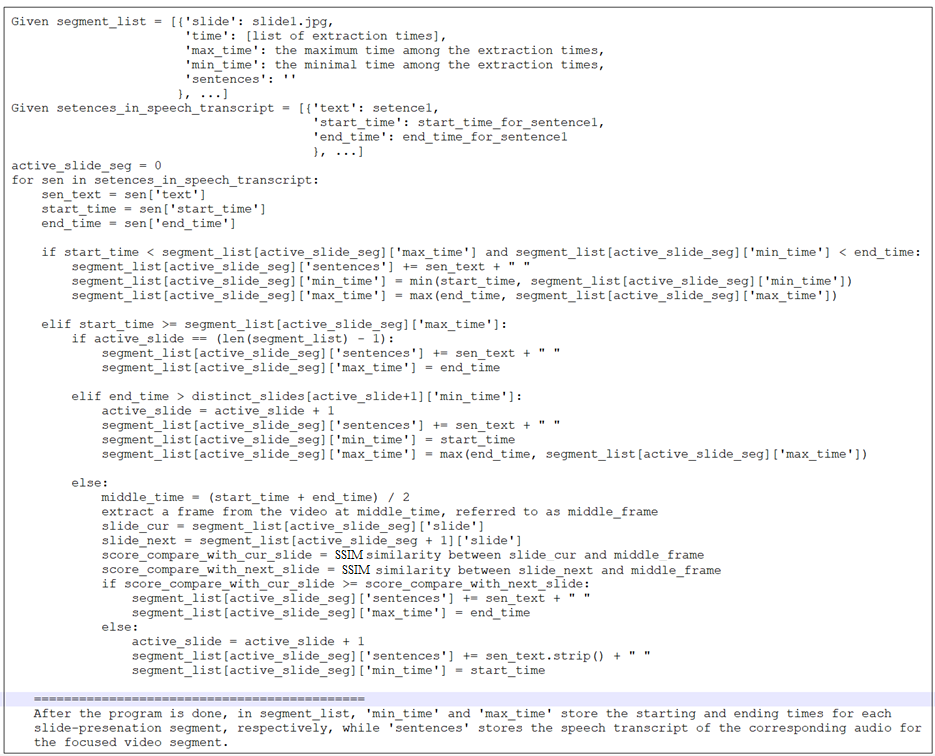}
    \caption{Detailed algorithm for $Step_B$ of the proposed Video Segmentation approach. In the given segment list, the list of extraction times refer to the times of the sampled frames that are deemed to contain the same slide in $Step_A$.}
    \label{fig:video-seg-Step_B}
\end{figure*}

Figure~\ref{fig:video_segmentation_algorithm} shows the details of video (scene) segmentation algorithm, which leverages off-the-shelf tool PySceneDetect. 
Meanwhile, a more detailed description of pseudo code is shown in Figure~\ref{fig:video-seg-Step_B} for reference.

In the proposed segmentation algorithm, we merge short segments (e.g., less than 3s) with the proceeding ones, assuming they are transitions between slides.  For the remaining segments with reasonable duration (e.g., one minute or less), we re-check whether the current segment actually presents the same slide as the previous segment, first using efficient SSIM (structural similarity index) \cite{1284395} to identify obvious cases and then using VLM to verify the remaining tricky ones.  If the answer is yes, the two segments are merged.  For the relatively long segments (i.e., $duration >$ one minute), we suspect the segment may actually contain multiple similar slides, and thus re-detect slides ($Step_A$) using VLM and determine the presentation time span for each detected slide ($Step_B$) based on vision/audio hints in this segment.  In $Step_A$, we sample a video frame every $N\_sample$ seconds in the focused segment, and use VLM to compare each frame with the previous one to check whether they contain a same slide.  Whenever the answer is negative, a new slide is identified. In $Step_B$, we leverage Automatic Speech Recognition (ASR) results (i.e., sentences with time stamps) of the focused segment and sample a video frame from the middle point of the time span for each sentence.  By comparing the extracted frame per sentence with its neighborhood detected slides with image similarity, we can determine which slide the sentence explains, and thus in this way, estimate the presentation time span of each detected slide in the focused segment.  The detailed approach of $Step_B$ is described in Figure~\ref{fig:video-seg-Step_B}. Throughout the segmentation algorithm, we use a same VLM and the same prompt (shown in Table~\ref{tab:prompt-table}) to determine whether two frames contain a same slide.

\subsection{LPM videos for Video Segmentation}
\label{sec:appendix_lpm_videos}
Links for the LPM videos used for video segmentation evaluation can be found in Table~\ref{tab:lpm-videos-link-table}. Note only the first 10 minutes are used.\newline
\input{latex/LPM_segmentation_evaluation_videos}

\subsection{Benchmark datasets Creation from Human Annotation.}
\label{sec:appendix_human_annotation}
We take the following measures to ensure the quality of MTurk annotation.

\begin{figure*}[b]
    \centering
    \includegraphics[scale=0.75]{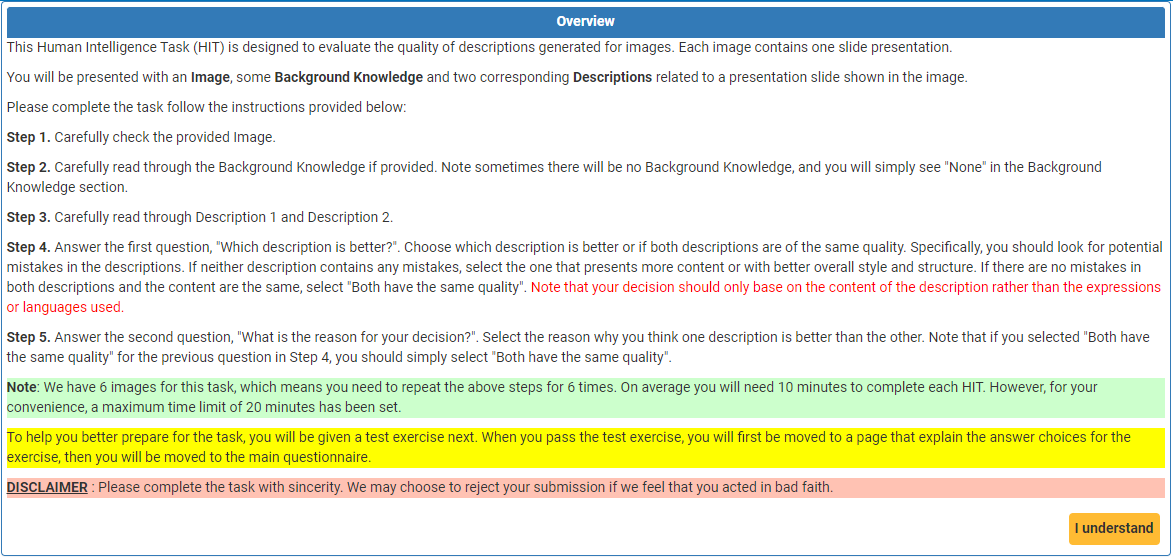}
    \caption{MTurk questionnaire instructions page.}
    \label{fig:questionnaire_intro}
\end{figure*}

\begin{figure*}[t!]
    \centering
    \includegraphics[scale=0.75]{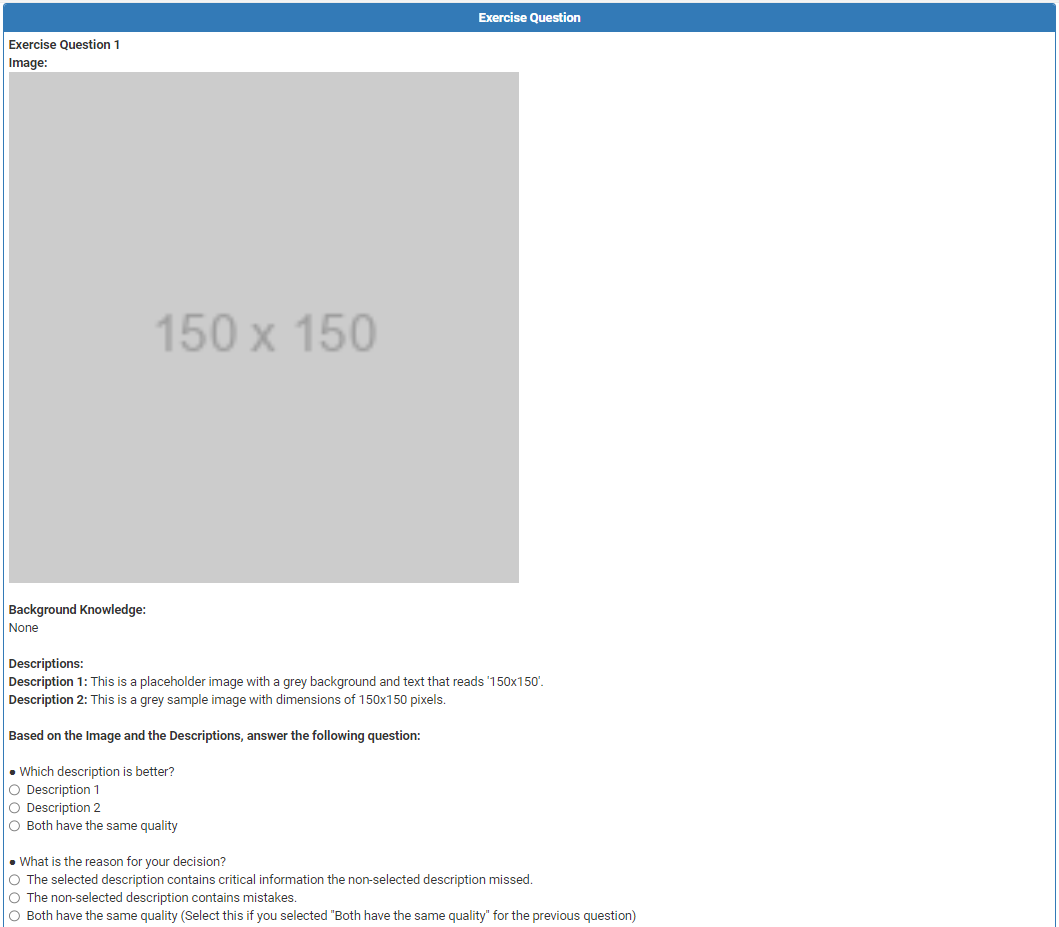}
    \caption{MTurk questionnaire exercise question 1.}
    \label{fig:questionnaire_exercise1}
\end{figure*}

\begin{figure*}[t!]
    \centering
    \includegraphics[scale=0.75, trim={1.5cm 0cm 1.5cm 1cm}]{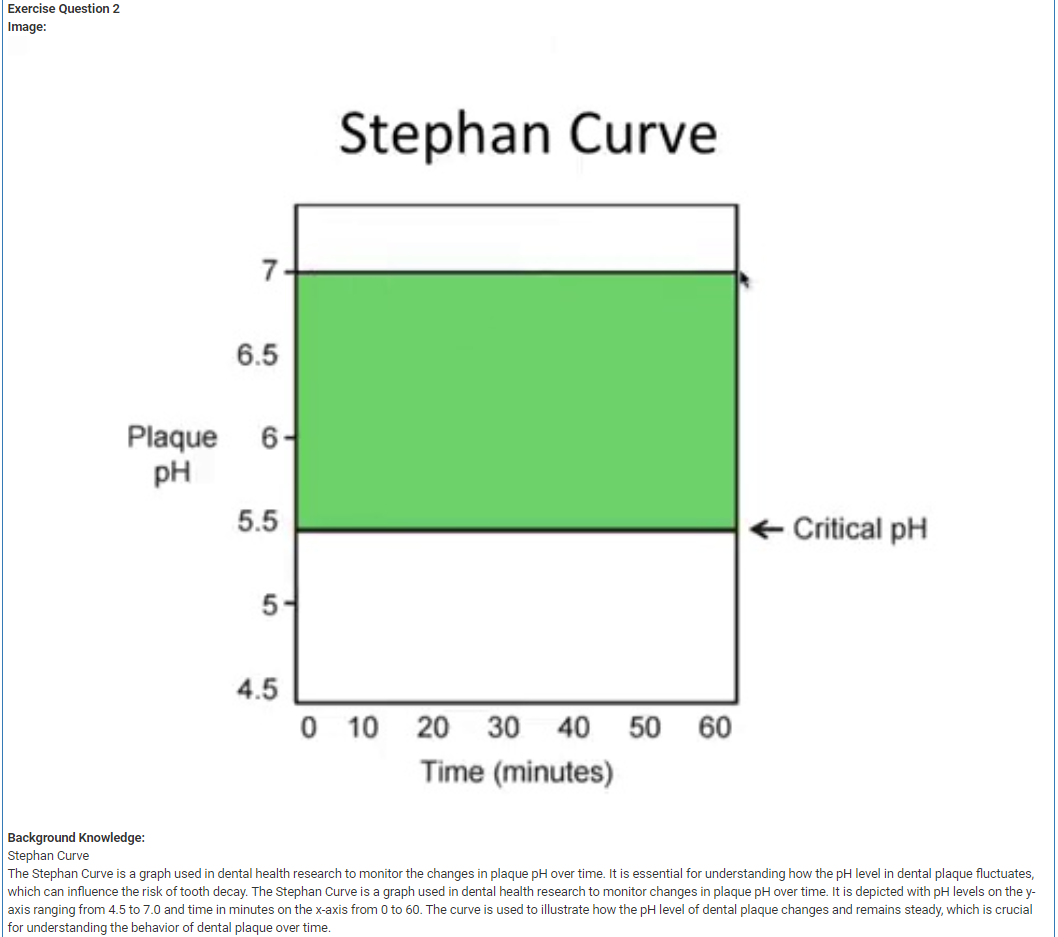}
    \includegraphics[scale=0.75, trim={1.5cm 0cm 1.5cm 0cm}]{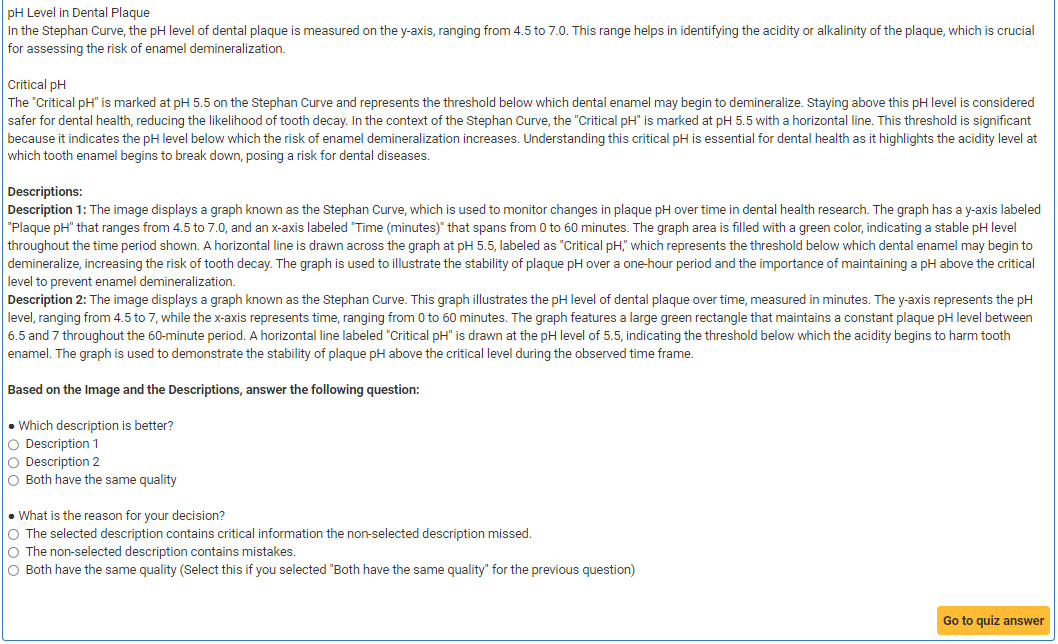}
    \caption{MTurk questionnaire exercise question 2. Note that the questionnaires that need annotation share the same format as this exercise question.}
    \label{fig:questionnaire_exercise2}
\end{figure*}

\begin{figure*}[t!]
    \centering
    \includegraphics[scale=0.75]{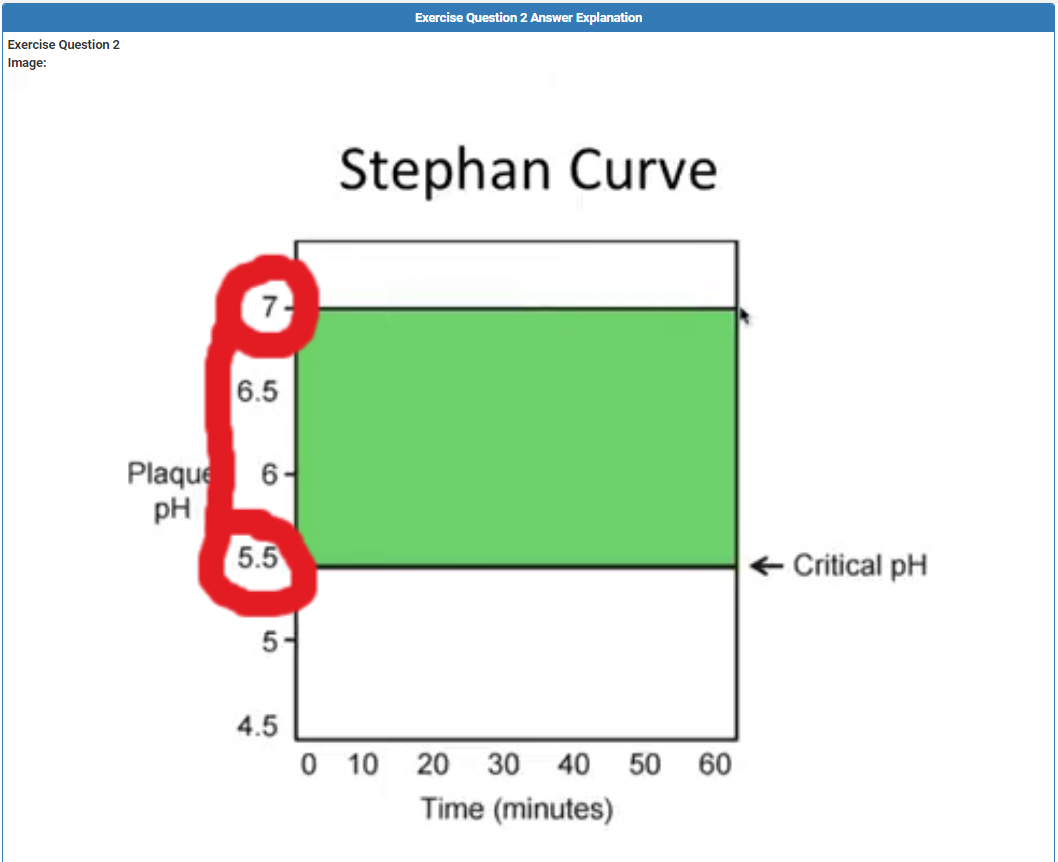}
    \includegraphics[scale=0.75, trim={1.5cm 0cm 1.5cm 1cm}]{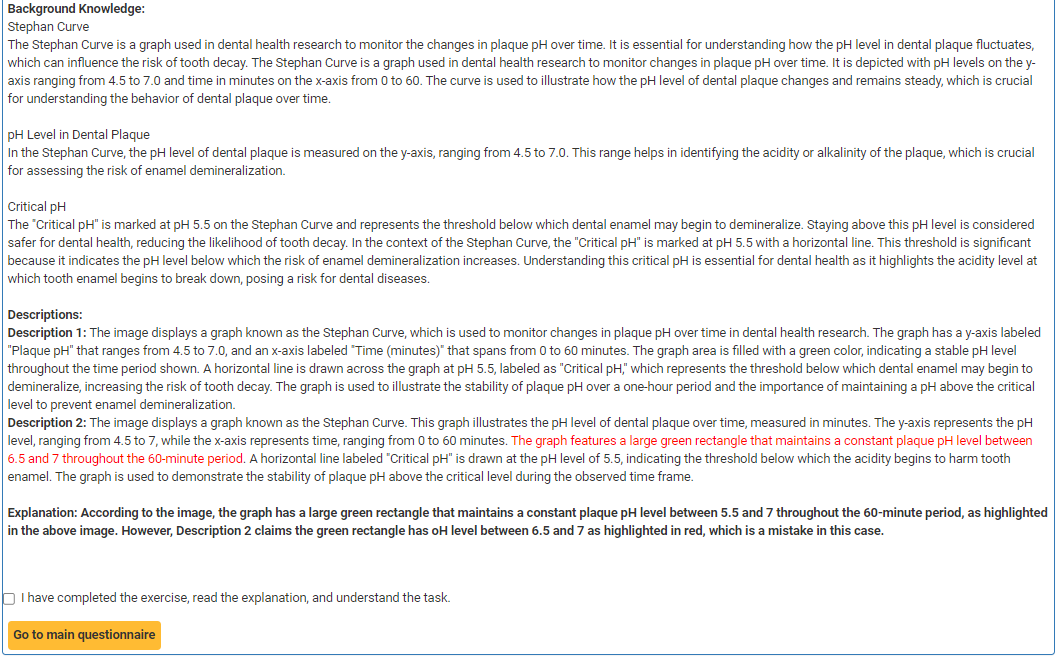}
    \caption{Answer explanation for MTurk questionnaire exercise question 2.}
    \label{fig:questionnaire_exercise2_explanations}
\end{figure*}

\begin{itemize}
    \item \textbf{Workers qualification: } We recruit MTurk workers who have ``Master qualification'', a Life-Time-Approval-Rate of at least $90\%$, and at least $10,000$ tasks approved. In addition, workers are preferred if English is their first language, as some lecture videos might be difficult to follow.
    \item \textbf{Test Exercises and Dummy Questions: } Workers are asked to complete two exercises, which are similar to a typical questionnaire, before working on a questionnaire set. Figure~\ref{fig:questionnaire_intro} is the starting page for the MTurkers, and as shown in Figure~\ref{fig:questionnaire_exercise1} and \ref{fig:questionnaire_exercise2} show the two exercises for MTurkers, so that they can understand the assigned task better. If the workers answer the exercises correctly, they will proceed to a page with explanations for the exercises, as shown in Figure~\ref{fig:questionnaire_exercise2_explanations}. After checking a box stating that they fully understand the task, workers will proceed to the questionnaire set. Among the six questionnaires in the set, one question is exactly the same with one exercise. This dummy question is used for quality control. We assume workers who wrongly answered the dummy question either rushed through the task or don't understand the task. We thus reject the corresponding questionnaires for quality purpose. 
    \item \textbf{Compensation: } In order to attract qualified workers to work on our questionnaires, the compensation for the annotators is set as an hourly rate of \$9, which is higher than the US federal minimum wage of \$7.25,
\end{itemize}

Please note, some of the figures in this section are put back in the document due to its screenshot size.

\subsection{System Settings}
\label{sec:appendix_system_settings}
\subsubsection{Experimental settings for the video segmentation component}
\label{subsec:appendix_system_settings_videp_segmentation}
We evaluate our proposed video segmentation approach and compare it with the SOTA PySceneDetect baseline on the datasets listed in Table~\ref{tab:data-stats-video-segmentaion-table}. Note that our approach also uses PySceneDetect to conduct first-round video segmentation.  In our approach, we especially tune the PySceneDetect to minimize the chance of missing slides in segmentation.  For fair comparison, for the baseline PySceneDetect, we tune it again to achieve its best overall performance.  We tune all the algorithm parameters on a separate held-out video set. In the proposed video-segmentation approach, $Threshold_1$, $Threshold_2$, $Threshold_3$ and $Threshold_4$ are set as 3 seconds, 60 seconds, 0.9, 0.65, respectively, and $N\_sample$ is set as 60 seconds. For PySceneDetect, AdaptiveDetector is used with adaptive\_threshold set as 1 and min\_content\_val set as 10.  For the baseline PySceneDetect, ContentDetector is adopted with threshold set as 12.  In this work, we adopt GPT 4 Vision as the VLM used in our proposed segmentation algorithm.  

We set temperature as 0 and max\_tokens as 800 for all GPT 4 models used in this work.

\subsubsection{Experimental settings for the video-segment understanding component} 
\label{subsec:appendix_system_settings_videp_understanding}
In this work, GPT-4 Turbo is used for all agents that require a VLM or LLM in processing in the Video-Segment Understanding component. 

In ASR recognition, the Whisper model is used to transcribe speech into text for each video. To further reduce hallucination, among the corrections suggested by the VLM, we only modify $transcript_i$ accordingly if the suggestion is "$term_A$ should be $term_B$" and the acoustic difference level between $term_A$ and $term_B$ (evaluated using PyPhonetics\footnote{\url{https://pypi.org/project/pyphonetics/}}) is less than 5, which means the two terms likely have similar pronunciations. For ASR correction, the acoustic difference level between two terms is evaluated using the RefinedSoundex.distance function of PyPhonetics. 

For the dynamic critic, $N_{max}$ is set to 10.

\subsubsection{Experimental Settings of the QA system used in the extrinsic evaluation}
\label{subsec:appendix_system_settings_qa}
We setup a Retrieval Augmented Generation (RAG) based QA system for extrinsic evaluation. 
The RAG based QA system is composed of a retriever based on FAISS\footnote{\url{https://github.com/facebookresearch/faiss}} and a reader using LangChain\footnote{\url{https://www.langchain.com/}}.
In the index-building process, the three multi-modal agents generates the understanding result respectively (including vision understanding result, speech transcript, and consolidation information). For each segment, the retriever builds up an embedding vector for that segmented scene(slide), which is added to the FAISS index. The embedding model is SentenceTransformer/all-MiniLM-L6-v2.  In the retrieval and QA phase, the reader wraps up the top 5 retrieval results as context, and sends the question together with the context to GPT 3.5 for answer generation. For evaluation, we ask GPT-4 Turbo to generate 2 questions answerable with vision information but not answerable with speech (denoted as vision questions), 2 questions answerable with speech but not answerable with vision information (denoted as speech questions), and 2 questions that can be best answered with the consolidated information (denoted as consolidation questions).  We also require the model to generate the ground-truth answer at the same time for each generated question to facilitate evaluation (See Table~\ref{tab:prompt-table2} for the prompt).  We then run the retrieval-based QA system to generate an answer for each question.  The correctness of the answer generated by QA system is also evaluated by GPT-4 Turbo using the prompt shown in Table~\ref{tab:prompt-table3}.

\subsection{Prompts used in the Experiments}
\label{sec:appendix_prompts}
We list the prompts used in the experiments in the later part of Appendix, due to its table size. 
Table~\ref{tab:prompt-table} lists the prompts used in all the agents of the proposed framework. 
Table~\ref{tab:prompt-table2} shows the prompts used to generate the questions together with the corresponding ground-truth answers for the extrinsic evaluation with QA, the ablation study on ASR correction, and the ablation study on vision understanding, respectively.  
Table~\ref{tab:prompt-table3} presents (1) the prompt that is used to determine whether two vision understanding results are inconsistent (i.e., containing conflicting information), and (2) the prompt that is used to determine whether an answer generated by the QA system is correct based on the question and the corresponding ground-truth. All prompts are carefully designed.

\input{latex/prompt_table}
\input{latex/prompt_table2}
\input{latex/prompt_table3}

\end{document}

%% file: latex/data_stats_segmentation.tex
\begin{table}[H]
\normalsize
\begin{center}
\scalebox{0.8}{\begin{tabular}{ l|ccc} 
 \toprule
Dataset & Video \# & Video Segments \# & \begin{tabular}[c]{@{}c@{}}Total Video\\Length(mins)\end{tabular}  \\
 \midrule
 LPM & $6$ & $54$ & $60.0$\\
 EI & $7$ & $37$ & $56.3$ \\
\bottomrule
\end{tabular}}
\end{center}
\vspace{-0.3cm}
\caption{Dataset statistics for video segmentation evaluation. Segments refer to ground-truth segments manually labeled.}
\vspace{-0.3cm}
\label{tab:data-stats-video-segmentaion-table}
\end{table}

%% file: latex/data_stats_understanding.tex
\begin{table}[H]
\normalsize
\begin{center}
\scalebox{0.8}{\begin{tabular}{ l|rrr} 
 \toprule
  Dataset & Video \# & Video Segments \# & \begin{tabular}[c]{@{}c@{}}Total Video\\Length(hours)\end{tabular} \\
 \midrule
 LPM & $188$ & $1366$ & $28.17$\\
 EI & $66$ & $264$ & $5.96$ \\
\bottomrule
\end{tabular}}
\end{center}
\vspace{-0.3cm}
\caption{Dataset statistics for evalution on understanding performance. Segments are manually labeled for LPM data but automatically detected for EI data.}
\vspace{-0.3cm}
\label{tab:data-stats-video-segment-understanding-table}
\end{table}

%% file: latex/segmentation_table.tex
\begin{table}[H]
\normalsize
\begin{center}
\scalebox{0.6}{\begin{tabular}{ l|rr|rr} 
 \toprule
 Dataset & \multicolumn{2}{c|}{LPM} & \multicolumn{2}{c}{EI}\\
 \midrule
 Algorithm & PySceneDetect & Ours & PySceneDetect & Ours\\
 \midrule
 Precision & 77.50 & 100.00 & 94.59 & 100.00 \\
 Recall & 88.33 & 96.55 & 80.00 & 100.00\\
 F1 & 82.56 & 98.24 & 86.69 & 100.00\\
 \midrule
 IoU & 0.74 & 0.80 & 0.63 & 0.91\\
\bottomrule
\end{tabular}}
\end{center}
\vspace{-0.4cm}
\caption{Video segmentation results.}
\vspace{-0.3cm}
\label{tab:segmentaion-table}
\end{table}

%% file: latex/human_eval_table.tex
\begin{table}[H]
\small
\begin{center}
\scalebox{0.625}{\begin{tabular}{ l|l|r|rrr|r|r} 
 \toprule
 & & Total \# & Win \# & Tie \# & Lose \# & Win \% & (Win+Tie) \%\\
 \midrule
 \multirow{3}{*}{LPM}& B+K vs B & $155$ & $76$ & $49$ & $30$ & $49.03\%$  & $80.65\%$\\
 & B+K+A vs B & $155$ & $91$ & $39$ & $25$ & $58.71\%$  & $83.9\%$\\
 & B+K+A+D vs B & $160$ & $94$ & $32$ & $34$ & $58.75\%$  & $78.75\%$\\
 \midrule
 \multirow{3}{*}{EI}& B+K vs B & $35$ & $16$ & $10$ & $9$ & $45.71\%$  & $74.29\%$\\
 & B+K+A vs B & $39$ & $18$ & $13$ & $8$ & $46.15\%$ & $79.49\%$\\
 & B+K+A+D vs B & $34$ & $16$ & $13$ & $5$ & $47.06\%$ & $85.29\%$ \\
\bottomrule
\end{tabular}}
\end{center}
\vspace{-0.3cm}
\caption{Comparison on vision understanding results with human evaluation. B refers to the baseline system; K refers to the knowledge-related enhancement mechanism; A refers to the ASR correction mechanism; D refers to the dynamic critic self-reflection mechanism.}
\label{tab:human-eval-table}
\vspace{-0.4cm}
\end{table}

%% file: latex/main_table.tex
\begin{table*}[h]
\small
\begin{center}
\scalebox{0.9}{\begin{tabular}{ l|r|rrrr||r|rrrr} 
 \toprule
 Question & LPM  &\multicolumn{4}{c||}{\textbf{ Accuracy on LPM} (given index type)} & EI & \multicolumn{4}{c}{\textbf{Accuracy on EI} (given index type)}\\
 
 Type & Question \# & All & Vision & Speech & Consolidation & Question \# & All & Vision & Speech & Consolidation \\
 \midrule 
 Vision & $2716$ & $78.57$ & $78.76$ & $33.59$ & $74.96$ & $392$ & $76.27$ & $75.77$ & $23.21$ & $68.62$\\ 
 Speech & $2706$ & $70.81$ & $49.04$ & $67.63$ & $68.92$ & $394$ & $82.74$ & $46.45$ & $87.06$ & $71.83$\\
 Consolidation & $2722$ & $86.41$ & $79.21$ & $66.13$ & $87.88$ & $394$ & $90.36$ & $79.19$ & $79.44$ & $88.83$\\
 \midrule
 \textbf{Overall} & $8144$ & $78.61$ & $69.03$ & $55.77$ & $77.27$ & $1180$ & $83.13$ & $67.11$ & $63.31$ & $76.44$\\
\bottomrule
\end{tabular}}
\end{center}
\vspace{-0.3cm}
\caption{Question-answering performance on LPM and DC data. }
\vspace{-0.4cm}
\label{tab:main-table}
\end{table*}

%% file: latex/speech_table_updated.tex
\begin{table}[t]
\small
\begin{center}
\scalebox{1.0}{\begin{tabular}{ l|l|rrrr} 
 \toprule
 & \textbf{LPM} & \textbf{EI}\\
 \midrule
 Baseline + Knowledge & $67.82$ & $70.59$ \\
 \hspace{0.15in}+ ASR correction & $73.91$ & $82.35$\\
\bottomrule
\end{tabular}}
\end{center}
\vspace{-0.3cm}
\caption{QA performance (Accuracy) given the question set for ASR-correction ablation study (870 questions for LPM and 102 questions for EI).}
\label{tab:speech-table}
\vspace{-0.3cm}
\end{table}

%% file: latex/vision_table_updated.tex
\begin{table}[t]
\small
\begin{center}
\scalebox{1.0}{\begin{tabular}{ l|l|rrrr} 
 \toprule
 & \textbf{LPM} & \textbf{EI}\\
 \midrule
 Baseline & $36.88$ & $39.40$ \\
 \hspace{0.05in}+ Knowledge & $46.86$ & $48.48$\\
 \hspace{0.10in}+ ASR Correction & $46.20$ & $51.52$ \\
 \hspace{0.15in}+ Dynamic Critic & $83.13$ & $69.70$ \\
\bottomrule
\end{tabular}}
\end{center}
\vspace{-0.3cm}
\caption{QA performance (Accuracy) given the question set for vision-understanding ablation Study (320 questions for LPM and 66 questions for EI).}
\vspace{-0.2cm}
\label{tab:vision-table}
\end{table}

%% file: latex/case_study.tex
\begin{table}[H]
\small
\begin{center}
\scalebox{0.75}{\begin{tabular}{l|l|p{6cm}} 
 \toprule
  \multirow{10}{*}{Knowledge} & LPM & The trends of the metabolic rate of endothermy and ectorthermy are described in reverse. With external knowledge of metabolic rate , the trends are correctly described.\\
 \cmidrule{2-3}
  & EI & The description of "ERP layer" wrongly includes certain stages of "PLM layer" due to confusing slide layout. With external knowledge of PLM and ERP leveraged, this problem is fixed.\\
  \midrule
 \multirow{5}{*}{ASR Correction} & LPM & A professor's name "Bisque" is correctly changed to "Bisk".\\
 \cmidrule{2-3}
  & EI & A domain specific term "Nexeed", previously recognized as "NextSeat", is corrected.\\
  \midrule
  \multirow{5}{*}{Dynamic Critic} & LPM & The starting point for a shaded area in a figure previously recognized at 6.5, is correctly fixed to 5.5. \\
  \cmidrule{2-3}
  & EI & A Venn diagram previously missing from the description is added.\\
 \bottomrule
\end{tabular}}
\end{center}
\vspace{-0.2cm}
\caption{Case study for improvement brought by knowledge-related enhancement, ASR correction, and dynamic critic.}
\vspace{-0.2cm}
\label{tab:case-study-table}
\end{table}

%% file: latex/efficiency_table.tex
\begin{table}[t]
\small
\begin{center}
\scalebox{0.8}{\begin{tabular}{ l|cc} 
 \toprule
 \textbf{Video Segmentation} & \textbf{\# API calls} & \textbf{Video Length (s)}\\
 \midrule
 EI Data & $\approx4$ & $483$ \\
 LPM Data & $\approx3$ & $600$\\
 \midrule
 \textbf{Video-Segment Understanding} & \textbf{\# API calls} & \textbf{Time Lapse (s)}\\
 \midrule
 Baseline & $2$ & $40$\\
 \hspace{0.05in}+ Knowledge & $4$ & $65$\\
 \hspace{0.10in}+ ASR Correction & $4$ & $65$\\
 \hspace{0.15in}+ Dynamic Critic & $\approx 9$ & $107$ \\
\bottomrule
\end{tabular}}
\end{center}
\vspace{-0.2cm}
\caption{
Efficiency of the \premind \space framework. For the video segmentation procedure, the average length of the videos and the average number of API calls needed to segment one video are listed. For video-segment understanding, the average number of API calls and the average total time for generating indexes per video segment are listed. ASR API call is \textbf{not included} as we only need to call ASR model once per video.}
\vspace{-0.2cm}
\label{tab:efficiency-table}
\end{table}

%% file: latex/LPM_segmentation_evaluation_videos.tex
\begin{table}[t!]
\small
\begin{center}
\scalebox{0.8}{\begin{tabular}{ l|p{7cm}} 
 \toprule
 Video \# & Video Link\\
 \midrule
 V1 & \url{https://www.youtube.com/watch?v=2NgUY8f1pa8}\\
 V2 & \url{https://www.youtube.com/watch?v=2_dZ5GBlRgU}\\
 V3 & \url{https://www.youtube.com/watch?v=_Awekr6-ilg}\\
 V4 & \url{https://www.youtube.com/watch?v=BsXUWddl-as}\\
 V5 & \url{https://www.youtube.com/watch?v=N75gvrZfO24}\\
 V6 & \url{https://www.youtube.com/watch?v=_Jw3DQ7_pxg}\\
\bottomrule
\end{tabular}}
\end{center}
\caption{LPM videos used for video segemntation evaluation.}
\label{tab:lpm-videos-link-table}
\end{table}

%% file: latex/prompt_table.tex
\begin{table*}[h]
\centering
\scalebox{0.55}{
\begin{tabular}{p{2.5cm}| p{20cm}} 
    \toprule
    \centering Agent/Algorithm & \centering Prompt \tabularnewline [0.5ex]
    \midrule
    \centering Video Segmentation & \{image 1\}\newline
    \{image 2\}\newline
    For the two images provided, if both images appear to be digitally corrupted or distorted, answer with "Yes. Both images are corrupted." and terminate the response. Otherwise, do the following:
    For the two images provided, each should show a person or several people presenting a slide. Is the slide shown in the first image the same as the one shown in the second image? Please check very carefully for different texts within the slides.
    Please start your answer with "Yes. " if the slides are the same and "No. " if the slides are different. Then give an explanation for your answer. 
    \tabularnewline [0.5ex]
    \midrule
    \centering Vision Understanding Agent (Baseline System) & \{image\}\newline
    Given the image provided, please follow the following rules to generate a description: \newline
    (1) If this image contains a slide that occupies at least half of the image, please describe the content of that slide in detail. In this case, when generating the description, please only focus on the slide's content, and ignore the slide's bottom part such as slide number, footnotes, company logo, etc. as well as other parts of the image.  If there are one or more humans presented in the image, please also ignore the humans and don't include them in the description in this case. \newline
    (2) Otherwise, if there is no significant slide in the image, please simply describe the image.
    \tabularnewline [0.5ex]
    \midrule
    \centering Vision Understanding Agent (Knowledge-Related Enhancement) & \{image\}\newline
    Given the image provided, we have the following background knowledge that are likely to be relevant: \newline
    \{retrieved knowledge\}\newline
    Based on the given image and the background knowledge, please follow the following rules to generate a description: \newline
    (1) If this image contains a slide that occupies at least half of the image, please describe the content of that slide in detail. In this case, when generating the description, please only focus on the slide's content, and ignore the slide's bottom part such as slide number, footnotes, company logo, etc. as well as other parts of the image.  If there are one or more humans presented in the image, please also ignore the humans and don't include them in the description in this case. \newline
    (2) Otherwise, if there is no significant slide in the image, please simple describe the image.
    \tabularnewline [0.5ex]
    \midrule
    \centering Vision-Audio Consolidation Agent & Given a video of someone presenting a slide, the text description of the slide (Part\_1), and the speech narrative of the presentation (Part\_2) are provided below.  Please consolidate the two parts into a nice overall description of the video content. \newline
    ----------------------------------------------\newline
    Part\_1. The text description of slide: \{vision understanding result\} \newline
    ----------------------------------------------\newline
    Part\_2. The speech narrative of the presentation: \{speech transcript\}
    \tabularnewline [0.5ex]
    \midrule
    \centering  Vision-based Keyword Extraction and ASR Error Correction Agent & \{image\} \newline
    Transcribed speech explanation for the image above: \{speech transcript\}\newline
    ------------------------------------\newline
    Given the provided image and its speech explanation transcript listed above, if the provided image contains a slide, extract the keywords (i.e., important words or phrases) from the slide. Then, check the transcribed speech explanation to see whether any keyword is misrecognized as other word or word sequence with similar pronunciation.  Please generate the response following the format below:\newline
    \newline
    List of keywords:\newline
    - keyword1\newline
    - keyword2\newline
    - keyword3\newline
    ...\newline
    (If a detected keyword contains ',' or ';' in middle, it should be split into multiple keywords. If no keyword is detected or the slide/image is empty, just leave the list of keywords empty.)\newline
    \newline
    Answer for whether certain keyword(s) is misrecognized:  Yes or No
    (if the answer is Yes, provide the following explanation:)\newline
    The term *** should be ****.\newline
    The term *** should be ****.\newline
    The term *** might be ****.\newline
    The term *** might be ****.\newline
    ...
    \tabularnewline [0.5ex]
    \midrule
    \centering Knowledge Extraction Agent & Given the text description listed below, summarize the concepts presented in this text description. If the text description is not about a slide presentation, reply with 'No concept extracted'. When generating the output, please follow the following format:\newline
    Concept: Concept name\newline
    Knowledge of Concept: explanation..\newline
    -------------\newline
    Concept: Concept name\newline
    Knowledge of Concept: explanation..\newline
    -------------\newline
    Concept: Concept name\newline
    Knowledge of Concept: explanation..\newline
    -------------\newline
    ....\newline
    -------------\newline
    Text description:\{consolidated understanding result\}
    \tabularnewline [0.5ex]
    \midrule
    \centering Dynamic Critic & Critic Agent: Given an image that contains a slide presentation and a description about the slide presentation, decide whether the description can be further improved.
    If the description is not comprehensive or containing potential mistakes, ask Vision Understanding Agent to improve the description.
    Otherwise, if the description is comprehensive and accurate, DO NOT repeat the description and just reply 'TERMINATE!!!' to Admin. \newline
    ***************************************************\newline
    Vision Understanding Agent: You can generate detailed description of slide presentation based on image provided with previous knowledge. Start your response with 'Vision Understanding Result:'.\newline
    ***************************************************\newline
    User Proxy: \{image\} \newline
    Given the image provided, we have the following background knowledge that might be relevant: \{retrieved knowledge\}\newline
    Based on the given image and the background knowledge, please use the following rules to generate a description: \newline
    (1) If this image contains a slide that occupies at least half of the image, please describe the content of that slide in detail. In this case, when generating the description, please only focus on the slide's content, and ignore the slide's bottom part such as slide number, footnotes, company logo, etc. as well as other parts of the image.  If there are one or more humans presented in the image, please also ignore the humans and don't include them in the description in this case. \newline
    (2) Otherwise, if the image DOES NOT contain a slide that occupies at least half of the image, just reply 'TERMINATE!!!' to Admin.
    \tabularnewline [0.5ex]
    \bottomrule
\end{tabular}}
\caption{Prompts for all agents used in this paper.}
\label{tab:prompt-table}
\end{table*}

%% file: latex/prompt_table2.tex
\begin{table*}[h]
\centering
\scalebox{0.5}{
\begin{tabular}{p{2.5cm}| p{21cm}} 
    \toprule
    \centering Function & \centering Prompt 
    \tabularnewline [0.5ex]
    \midrule
    \centering Question Generation (extrinsic evaluation with QA)& We have a video segment in which a speaker is presenting a slide.  The information about the video segement is summarizied into three parts, listed below after the "----" line. The first part, denoted as "Part\_1. The text description of slide:", provides a text description of the vision information presented in the slide. The second part, denoted as "Part\_2. The speech narrative of the presentation:", provides the speech-to-text transcript about what the speaker said about the slide.  The third part, denoted as "Part\_3. Info-Consolidation Output:", provides the overall description of the video segment, which consolidates the information from the first part and second part.\newline
    Your task is to generate six questions satisfying following requirements respectively, and also generate the corresponding answer for each generated question. \newline
    (1) Generate two questions (referred to as Question\_1\_vision and Question\_2\_vision) that are answerable with the slide vision information (i.e., information provided in the first part), but not answerable with the speech transcript (i.e., information provided in the second part). \newline
    (2) Generate two questions (referred to as Question\_3\_speech and Question\_4\_speech) that are answerable with the speech transcript (i.e., information provided in the second part), but not answerable with the slide vision information (i.e., information provided in the first part). \newline
    (3) Generate two questions (referred to as Question\_5\_consolidated and Question\_6\_consolidated) that can be best answered with the consolidated information, i.e., information provided in the third part. \newline
    
    In addition, when generting a question or a answer, please directly talk about the knowledge point and avoid the mentioning of slide, speech/speaker, and video as information sources (e.g., avoid the use of "according to the speech", "according to the presentation", "according to according to the consolidated information", "as discussed in the video", "according to the speaker", etc.).  For example, instead of asking a question "How does Business Intelligence aid in the preparation of decisions according to the presentation?", please directly ask "How does Business Intelligence aid in the preparation of decisions?".  Another example is that instead of answering "The slide indicates that Business Intelligence helps interpret past data to inform future decisions.", please directly answer "Business Intelligence helps interpret past data to inform future decisions.".\newline
    Please also note that if "slide" must be mentioned in a question, please always include slide title to show which slide it is about.\newline
    Please provide the generated questions and the answers in the following format:\newline
    
    Question\_1\_vision:\newline
    Answer\_1\_vision:\newline
    Question\_2\_vision:\newline
    Answer\_2\_vision:\newline
    Question\_3\_speech:\newline
    Answer\_3\_speech:\newline
    Question\_4\_speech:\newline
    Answer\_4\_speech:\newline
    Question\_5\_consolidated:\newline
    Answer\_5\_consolidated:\newline
    Question\_6\_consolidated:\newline
    Answer\_6\_consolidated:\newline
    ----------------------------------------------\newline
    Part 1. The text description of slide: \{vision understanding result\}\newline
    ----------------------------------------------\newline
    Part 2. The speech narrative of the presentation: \{speech transcript\}\newline
    ----------------------------------------------\newline
    Part 3. Info-Consolidation Output: \{consolidated information\}
    \tabularnewline [0.5ex]
    \midrule
    \centering Question Generation (ablation study on vision understanding) & We have a video segment in which a speaker is presenting a slide. Two visual descriptions about the video segment and their differences are provided.\newline
    Your task is to generate two questions satisfying following requirements, and also generate the corresponding answer for each generated question. \newline
    Generate two questions (referred to as Question\_1\_vision and Question\_2\_vision) that are answerable with Description 2, but not answerable with Description 1. \newline
    
    In addition, when generating a question or a answer, please directly talk about the knowledge point and avoid the mentioning of slide, speech/speaker, and video as information sources (e.g., avoid the use of "according to the speech", "according to the presentation", "according to according to the consolidated information", "as discussed in the video", "according to the speaker", etc.).  For example, instead of asking a question "How does Business Intelligence aid in the preparation of decisions according to the presentation?", please directly ask "How does Business Intelligence aid in the preparation of decisions?".  Another example is that instead of answering "The slide indicates that Business Intelligence helps interpret past data to inform future decisions.", please directly answer "Business Intelligence helps interpret past data to inform future decisions.".\newline
    Please also note that if "slide" must be mentioned in a question, please always include slide title to show which slide it is about.\newline
    Please provide the generated questions and the answers in the following example format:\newline
    
    Question\_1\_vision:\newline
    Answer\_1\_vision:\newline
    Question\_2\_vision:\newline
    Answer\_2\_vision:\newline
    ----------------------------------------------\newline
    Description 1: \{vision understanding result 1\}\newline
    ----------------------------------------------\newline
    Description 2: \{vision understanding result 2\}\newline
    ----------------------------------------------\newline
    Difference detection result: \{difference\}
    \tabularnewline [0.5ex]
    \midrule
    \centering Question Generation (ablation study on ASR correction) & We have a video segment in which a speaker is presenting a slide. The transcript about the video segment is provided and the corrections, if any, needed for the transcript.\newline
    Your task is to generate two questions satisfying following requirements, and also generate the corresponding answer for each generated question. \newline
    Generate two questions (referred to as Question\_3\_speech and Question\_4\_speech) that are answerable with the speech transcript correction, but not answerable without the speech transcript correction.\newline
    
    In addition, when generating a question or a answer, please directly talk about the knowledge point and avoid the mentioning of slide, speech/speaker, and video as information sources (e.g., avoid the use of "according to the speech", "according to the presentation", "according to according to the consolidated information", "as discussed in the video", "according to the speaker", etc.).  For example, instead of asking a question "How does Business Intelligence aid in the preparation of decisions according to the presentation?", please directly ask "How does Business Intelligence aid in the preparation of decisions?".  Another example is that instead of answering "The slide indicates that Business Intelligence helps interpret past data to inform future decisions.", please directly answer "Business Intelligence helps interpret past data to inform future decisions.".\newline
    Please also note that if "slide" must be mentioned in a question, please always include slide title to show which slide it is about.\newline
    Please provide the generated questions and the answers in the following example format:\newline
    
    Question\_3\_speech:\newline
    Answer\_3\_speech:\newline
    Question\_4\_speech:\newline
    Answer\_4\_speech:\newline
    ----------------------------------------------\newline
    Transcript: \{original speech transcript\}\newline
    ----------------------------------------------\newline
    Transcript correction needed:: \{corrections\}
    \tabularnewline [0.5ex]
    \bottomrule
\end{tabular}}
\caption{Prompts for question generation.}
\label{tab:prompt-table2}
\end{table*}

%% file: latex/prompt_table3.tex
\begin{table*}[h]
\centering
\scalebox{0.6}{
\begin{tabular}{p{2.5cm}| p{15cm}} 
    \toprule
    \centering Function & \centering Prompt 
    \tabularnewline [0.5ex]
    \midrule
    \centering Inconsistency Detection & Given two descriptions about a same slide (listed below), please determine whether there is any conflict in meaning between the two descriptions. \newline
    Please first answer 'Yes' or 'No', and if the answer is 'Yes', explain what the meaning conflict(s) is? \newline
    ---------------------- \newline
    Description 1: \{vision understanding 1\}\newline
    ---------------------- \newline
    Description 2: \{vision understanding 2\}
    \tabularnewline [0.5ex]
    \midrule
    \centering QA Evaluation (Answer Correctness) & Given a question and its ground-truth answer, check whether a automatically generated answer is correct.  The question, ground-truth answer, and the automatically generated answer are all listed below.  In your response, please simply say "correct" if you think the generated answer contains consistent information as the ground-truth answer, simply say "wrong" if you think the generated answer is wrong (i.e., conflicting with the information in the ground-truth answer, or failing to include the key messages in the ground-truth answer), and simply say "correct but with additional information" if you think the generated answer contains the correct answer but includes additional information not mentioned in the ground-truth answer.\newline
    \newline
    Question: \{question\}\newline\newline
    Ground-truth answer: \{ground truth answer\}\newline\newline
    Automatically generated answer: \{predicted answer\}
    \tabularnewline [0.5ex]
    \bottomrule
\end{tabular}}
\caption{Prompts for inconsistency detection given two vision understanding results and for answer correctness evaluation.}
\label{tab:prompt-table3}
\end{table*}

%% file: acl_latex.bbl
\begin{thebibliography}{47}
\providecommand{\natexlab}[1]{#1}

\bibitem[{Abdullah et~al.(2024)Abdullah, Liu, Wei, Kong, and Huang}]{abdullah2024ualbenchcomprehensiveunusualactivity}
Hasnat~Md Abdullah, Tian Liu, Kangda Wei, Shu Kong, and Ruihong Huang. 2024.
\newblock \href {https://arxiv.org/abs/2410.01180} {Ual-bench: The first comprehensive unusual activity localization benchmark}.
\newblock \emph{Preprint}, arXiv:2410.01180.

\bibitem[{Arazzi et~al.(2023)Arazzi, Ferretti, and Nocera}]{Arazzi2023SemanticHI}
Marco Arazzi, Marco Ferretti, and Antonino Nocera. 2023.
\newblock \href {https://api.semanticscholar.org/CorpusID:259044832} {Semantic hierarchical indexing for online video lessons using natural language processing}.
\newblock \emph{Big Data Cogn. Comput.}, 7:107.

\bibitem[{Chand and Oğul(2021)}]{9378632}
Dipesh Chand and Hasan Oğul. 2021.
\newblock \href {https://doi.org/10.1109/SAMI50585.2021.9378632} {A framework for lecture video segmentation from extracted speech content}.
\newblock In \emph{2021 IEEE 19th World Symposium on Applied Machine Intelligence and Informatics (SAMI)}, pages 000299--000304.

\bibitem[{Chang et~al.(2024)Chang, Wang, Wang, Wu, Yang, Zhu, Chen, Yi, Wang, Wang et~al.}]{chang2024survey}
Yupeng Chang, Xu~Wang, Jindong Wang, Yuan Wu, Linyi Yang, Kaijie Zhu, Hao Chen, Xiaoyuan Yi, Cunxiang Wang, Yidong Wang, et~al. 2024.
\newblock A survey on evaluation of large language models.
\newblock \emph{ACM Transactions on Intelligent Systems and Technology}, 15(3):1--45.

\bibitem[{Chen et~al.(2023)Chen, Zheng, Wang, Xu, Huang, Pan, Wang, Wang, Qiao, Lu, and Wang}]{chen2023videollmmodelingvideosequence}
Guo Chen, Yin-Dong Zheng, Jiahao Wang, Jilan Xu, Yifei Huang, Junting Pan, Yi~Wang, Yali Wang, Yu~Qiao, Tong Lu, and Limin Wang. 2023.
\newblock \href {https://arxiv.org/abs/2305.13292} {Videollm: Modeling video sequence with large language models}.
\newblock \emph{Preprint}, arXiv:2305.13292.

\bibitem[{Debnath et~al.(2023)Debnath, Rao, and Das}]{Debnath2023AML}
Abhijit Debnath, K.~Sreenivasa Rao, and Partha~Pratim Das. 2023.
\newblock \href {https://api.semanticscholar.org/CorpusID:266533926} {A multi-modal lecture video indexing and retrieval framework with multi-scale residual attention network and multi-similarity computation}.
\newblock \emph{Signal Image Video Process.}, 18:1993--2006.

\bibitem[{Dosovitskiy et~al.(2020)Dosovitskiy, Beyer, Kolesnikov, Weissenborn, Zhai, Unterthiner, Dehghani, Minderer, Heigold, Gelly, Uszkoreit, and Houlsby}]{Dosovitskiy2020AnII}
Alexey Dosovitskiy, Lucas Beyer, Alexander Kolesnikov, Dirk Weissenborn, Xiaohua Zhai, Thomas Unterthiner, Mostafa Dehghani, Matthias Minderer, Georg Heigold, Sylvain Gelly, Jakob Uszkoreit, and Neil Houlsby. 2020.
\newblock \href {https://api.semanticscholar.org/CorpusID:225039882} {An image is worth 16x16 words: Transformers for image recognition at scale}.
\newblock \emph{ArXiv}, abs/2010.11929.

\bibitem[{Gruzman and Kostenkova(2014)}]{gruzman2014algorithm}
Igor~S Gruzman and Anna~S Kostenkova. 2014.
\newblock Algorithm of scene change detection in a video sequence based on the threedimensional histogram of color images.
\newblock In \emph{2014 12th International Conference on Actual Problems of Electronics Instrument Engineering (APEIE)}, pages 1--1. IEEE.

\bibitem[{Hatalis et~al.(2024)Hatalis, Christou, Myers, Jones, Lambert, Amos-Binks, Dannenhauer, and Dannenhauer}]{Hatalis_Christou_Myers_Jones_Lambert_Amos-Binks_Dannenhauer_Dannenhauer_2024}
Kostas Hatalis, Despina Christou, Joshua Myers, Steven Jones, Keith Lambert, Adam Amos-Binks, Zohreh Dannenhauer, and Dustin Dannenhauer. 2024.
\newblock \href {https://doi.org/10.1609/aaaiss.v2i1.27688} {Memory matters: The need to improve long-term memory in llm-agents}.
\newblock \emph{Proceedings of the AAAI Symposium Series}, 2(1):277--280.

\bibitem[{Huang et~al.(2024)Huang, Wang, Chen, Song, and Zhu}]{Huang_2024_CVPR}
Bin Huang, Xin Wang, Hong Chen, Zihan Song, and Wenwu Zhu. 2024.
\newblock Vtimellm: Empower llm to grasp video moments.
\newblock In \emph{Proceedings of the IEEE/CVF Conference on Computer Vision and Pattern Recognition (CVPR)}, pages 14271--14280.

\bibitem[{Ip and Chan(1997)}]{10.1145/267437.267478}
Horace Ho-Shing Ip and Siu-Lok Chan. 1997.
\newblock \href {https://doi.org/10.1145/267437.267478} {Hypertext-assisted video indexing and content-based retrieval}.
\newblock In \emph{Proceedings of the Eighth ACM Conference on Hypertext}, HYPERTEXT '97, page 232–233, New York, NY, USA. Association for Computing Machinery.

\bibitem[{Iyer et~al.(2019)Iyer, Parekh, Mohandoss, Ramsurat, Raj, and Singh}]{iyer2019contentbasedvideoindexingretrieval}
Rahul~Radhakrishnan Iyer, Sanjeel Parekh, Vikas Mohandoss, Anush Ramsurat, Bhiksha Raj, and Rita Singh. 2019.
\newblock \href {https://arxiv.org/abs/1602.08581} {Content-based video indexing and retrieval using corr-lda}.
\newblock \emph{Preprint}, arXiv:1602.08581.

\bibitem[{Jeong et~al.(2012)Jeong, Kim, and Kim}]{Jeong2012AnAL}
Hyun~Ji Jeong, Tak-Eun Kim, and Myoung-Ho Kim. 2012.
\newblock \href {https://api.semanticscholar.org/CorpusID:14262991} {An accurate lecture video segmentation method by using sift and adaptive threshold}.
\newblock In \emph{Advances in Mobile Multimedia}.

\bibitem[{Lee et~al.(2023)Lee, Ahuja, Liang, Natu, and Morency}]{Lee_2023_ICCV}
Dong~Won Lee, Chaitanya Ahuja, Paul~Pu Liang, Sanika Natu, and Louis-Philippe Morency. 2023.
\newblock Lecture presentations multimodal dataset: Towards understanding multimodality in educational videos.
\newblock In \emph{Proceedings of the IEEE/CVF International Conference on Computer Vision (ICCV)}, pages 20087--20098.

\bibitem[{Li et~al.(2024)Li, Xu, Zhang, Song, Cai, Qi, Zhou, Pan, Li, Tu, Huang, and Wang}]{li-etal-2024-groundinggpt}
Zhaowei Li, Qi~Xu, Dong Zhang, Hang Song, YiQing Cai, Qi~Qi, Ran Zhou, Junting Pan, Zefeng Li, Vu~Tu, Zhida Huang, and Tao Wang. 2024.
\newblock \href {https://doi.org/10.18653/v1/2024.acl-long.360} {{G}rounding{GPT}: Language enhanced multi-modal grounding model}.
\newblock In \emph{Proceedings of the 62nd Annual Meeting of the Association for Computational Linguistics (Volume 1: Long Papers)}, pages 6657--6678, Bangkok, Thailand. Association for Computational Linguistics.

\bibitem[{Liang et~al.(2024)Liang, He, Jiao, Wang, Wang, Wang, Yang, Shi, and Tu}]{liang-etal-2024-encouraging}
Tian Liang, Zhiwei He, Wenxiang Jiao, Xing Wang, Yan Wang, Rui Wang, Yujiu Yang, Shuming Shi, and Zhaopeng Tu. 2024.
\newblock \href {https://doi.org/10.18653/v1/2024.emnlp-main.992} {Encouraging divergent thinking in large language models through multi-agent debate}.
\newblock In \emph{Proceedings of the 2024 Conference on Empirical Methods in Natural Language Processing}, pages 17889--17904, Miami, Florida, USA. Association for Computational Linguistics.

\bibitem[{Lin et~al.(2023)Lin, Ye, Zhu, Cui, Ning, Jin, and Yuan}]{lin2023videollavalearningunitedvisual}
Bin Lin, Yang Ye, Bin Zhu, Jiaxi Cui, Munan Ning, Peng Jin, and Li~Yuan. 2023.
\newblock \href {https://arxiv.org/abs/2311.10122} {Video-llava: Learning united visual representation by alignment before projection}.
\newblock \emph{Preprint}, arXiv:2311.10122.

\bibitem[{Lin et~al.(2024)Lin, Ye, Zhu, Cui, Ning, Jin, and Yuan}]{lin-etal-2024-video}
Bin Lin, Yang Ye, Bin Zhu, Jiaxi Cui, Munan Ning, Peng Jin, and Li~Yuan. 2024.
\newblock \href {https://doi.org/10.18653/v1/2024.emnlp-main.342} {Video-{LL}a{VA}: Learning united visual representation by alignment before projection}.
\newblock In \emph{Proceedings of the 2024 Conference on Empirical Methods in Natural Language Processing}, pages 5971--5984, Miami, Florida, USA. Association for Computational Linguistics.

\bibitem[{Lin et~al.(2004)Lin, Nunamaker, Chau, and Chen}]{1265045}
Ming Lin, J.F. Nunamaker, M.~Chau, and Hsinchun Chen. 2004.
\newblock \href {https://doi.org/10.1109/HICSS.2004.1265045} {Segmentation of lecture videos based on text: a method combining multiple linguistic features}.
\newblock In \emph{37th Annual Hawaii International Conference on System Sciences, 2004. Proceedings of the}, pages 9 pp.--.

\bibitem[{Ma et~al.(2017)Ma, Zhang, Ouyang, and Agam}]{8260637}
Di~Ma, Xi~Zhang, Xu~Ouyang, and Gady Agam. 2017.
\newblock \href {https://doi.org/10.1109/ICMLA.2017.0-155} {Lecture vdeo indexing using boosted margin maximizing neural networks}.
\newblock In \emph{2017 16th IEEE International Conference on Machine Learning and Applications (ICMLA)}, pages 221--227.

\bibitem[{Ma et~al.(2024)Ma, Jin, Wang, Xian, Feng, and Yang}]{Ma_2024_CVPR}
Fan Ma, Xiaojie Jin, Heng Wang, Yuchen Xian, Jiashi Feng, and Yi~Yang. 2024.
\newblock Vista-llama: Reducing hallucination in video language models via equal distance to visual tokens.
\newblock In \emph{Proceedings of the IEEE/CVF Conference on Computer Vision and Pattern Recognition (CVPR)}, pages 13151--13160.

\bibitem[{Ma et~al.(2025)Ma, Qian, Gales, and Knill}]{ma2025asrerrorcorrectionusing}
Rao Ma, Mengjie Qian, Mark Gales, and Kate Knill. 2025.
\newblock \href {https://arxiv.org/abs/2409.09554} {Asr error correction using large language models}.
\newblock \emph{Preprint}, arXiv:2409.09554.

\bibitem[{Ma et~al.(2023)Ma, Qian, Manakul, Gales, and Knill}]{ma2023generativelargelanguagemodels}
Rao Ma, Mengjie Qian, Potsawee Manakul, Mark Gales, and Kate Knill. 2023.
\newblock \href {https://arxiv.org/abs/2307.04172} {Can generative large language models perform asr error correction?}
\newblock \emph{Preprint}, arXiv:2307.04172.

\bibitem[{Maaz et~al.(2024)Maaz, Rasheed, Khan, and Khan}]{maaz-etal-2024-video}
Muhammad Maaz, Hanoona Rasheed, Salman Khan, and Fahad Khan. 2024.
\newblock \href {https://doi.org/10.18653/v1/2024.acl-long.679} {Video-{C}hat{GPT}: Towards detailed video understanding via large vision and language models}.
\newblock In \emph{Proceedings of the 62nd Annual Meeting of the Association for Computational Linguistics (Volume 1: Long Papers)}, pages 12585--12602, Bangkok, Thailand. Association for Computational Linguistics.

\bibitem[{Madaan et~al.(2023)Madaan, Tandon, Gupta, Hallinan, Gao, Wiegreffe, Alon, Dziri, Prabhumoye, Yang, Gupta, Majumder, Hermann, Welleck, Yazdanbakhsh, and Clark}]{madaan2023selfrefineiterativerefinementselffeedback}
Aman Madaan, Niket Tandon, Prakhar Gupta, Skyler Hallinan, Luyu Gao, Sarah Wiegreffe, Uri Alon, Nouha Dziri, Shrimai Prabhumoye, Yiming Yang, Shashank Gupta, Bodhisattwa~Prasad Majumder, Katherine Hermann, Sean Welleck, Amir Yazdanbakhsh, and Peter Clark. 2023.
\newblock \href {https://arxiv.org/abs/2303.17651} {Self-refine: Iterative refinement with self-feedback}.
\newblock \emph{Preprint}, arXiv:2303.17651.

\bibitem[{Medida and KASARAPU(2021)}]{medida2021video}
Lakshmi Medida and RAMANI KASARAPU. 2021.
\newblock \href {https://doi.org/10.22937/IJCSNS.2021.21.8.12} {An optimized e-lecture video search and indexing framework}.
\newblock 21:87--96.

\bibitem[{Mishra et~al.(2023)Mishra, Raj, Kumar, Kasaudhan, Kumar~Mishra, and Maini}]{10170589}
Gouri~Shankar Mishra, Anand Raj, Amit Kumar, Aman~Kumar Kasaudhan, Pradeep Kumar~Mishra, and Tarun Maini. 2023.
\newblock \href {https://doi.org/10.1109/INCET57972.2023.10170589} {Indexing and segmentation of video contents: A review}.
\newblock In \emph{2023 4th International Conference for Emerging Technology (INCET)}, pages 1--9.

\bibitem[{Pan et~al.(2023)Pan, Lin, Ge, Zhu, Zhang, Wang, Qiao, and Li}]{pan2023retrieving}
Junting Pan, Ziyi Lin, Yuying Ge, Xiatian Zhu, Renrui Zhang, Yi~Wang, Yu~Qiao, and Hongsheng Li. 2023.
\newblock Retrieving-to-answer: Zero-shot video question answering with frozen large language models.
\newblock In \emph{Proceedings of the IEEE/CVF International Conference on Computer Vision}, pages 272--283.

\bibitem[{Radford et~al.(2021)Radford, Kim, Hallacy, Ramesh, Goh, Agarwal, Sastry, Askell, Mishkin, Clark, Krueger, and Sutskever}]{radford2021learningtransferablevisualmodels}
Alec Radford, Jong~Wook Kim, Chris Hallacy, Aditya Ramesh, Gabriel Goh, Sandhini Agarwal, Girish Sastry, Amanda Askell, Pamela Mishkin, Jack Clark, Gretchen Krueger, and Ilya Sutskever. 2021.
\newblock \href {https://arxiv.org/abs/2103.00020} {Learning transferable visual models from natural language supervision}.
\newblock \emph{Preprint}, arXiv:2103.00020.

\bibitem[{Reddy and Jadhav(2015)}]{reddy2015comparison}
Bindu Reddy and Anita Jadhav. 2015.
\newblock Comparison of scene change detection algorithms for videos.
\newblock In \emph{2015 Fifth International Conference on Advanced Computing \& Communication Technologies}, pages 84--89. IEEE.

\bibitem[{Ren et~al.(2024)Ren, Yao, Li, Sun, and Hou}]{Ren_2024_CVPR}
Shuhuai Ren, Linli Yao, Shicheng Li, Xu~Sun, and Lu~Hou. 2024.
\newblock Timechat: A time-sensitive multimodal large language model for long video understanding.
\newblock In \emph{Proceedings of the IEEE/CVF Conference on Computer Vision and Pattern Recognition (CVPR)}, pages 14313--14323.

\bibitem[{Saoudi and Jai~Andaloussi(2021)}]{unknown}
ElMehdi Saoudi and Said Jai~Andaloussi. 2021.
\newblock \href {https://doi.org/10.21203/rs.3.rs-255106/v1} {A distributed content-based video retrieval system for large data-sets}.

\bibitem[{Shah et~al.(2015)Shah, Yu, Shaikh, and Zimmermann}]{Shah2015TRACELA}
Rajiv~Ratn Shah, Yi~Yu, Anwar~Dilawar Shaikh, and Roger Zimmermann. 2015.
\newblock \href {https://api.semanticscholar.org/CorpusID:38213200} {Trace: Linguistic-based approach for automatic lecture video segmentation leveraging wikipedia texts}.
\newblock \emph{2015 IEEE International Symposium on Multimedia (ISM)}, pages 217--220.

\bibitem[{Song et~al.(2024)Song, Chai, Wang, Zhang, Zhou, Wu, Chi, Guo, Ye, Zhang, Lu, Hwang, and Wang}]{song2024moviechatdensetokensparse}
Enxin Song, Wenhao Chai, Guanhong Wang, Yucheng Zhang, Haoyang Zhou, Feiyang Wu, Haozhe Chi, Xun Guo, Tian Ye, Yanting Zhang, Yan Lu, Jenq-Neng Hwang, and Gaoang Wang. 2024.
\newblock \href {https://arxiv.org/abs/2307.16449} {Moviechat: From dense token to sparse memory for long video understanding}.
\newblock \emph{Preprint}, arXiv:2307.16449.

\bibitem[{Soni and Dubey(2019)}]{Soni}
Shraddha Soni and Shubham Dubey. 2019.
\newblock \href {https://doi.org/10.32628/IJSRCSEIT} {Towards systematic literature review of e-learning}.

\bibitem[{Team et~al.(2024)Team, Anil, Borgeaud, Alayrac, Yu, Soricut, Schalkwyk, Dai, Hauth, and et~al}]{geminiteam2024geminifamilyhighlycapable}
Gemini Team, Rohan Anil, Sebastian Borgeaud, Jean-Baptiste Alayrac, Jiahui Yu, Radu Soricut, Johan Schalkwyk, Andrew~M. Dai, Anja Hauth, and et~al. 2024.
\newblock \href {https://arxiv.org/abs/2312.11805} {Gemini: A family of highly capable multimodal models}.
\newblock \emph{Preprint}, arXiv:2312.11805.

\bibitem[{Wang et~al.(2024{\natexlab{a}})Wang, Ma, Feng, Zhang, Yang, Zhang, Chen, Tang, Chen, Lin, Zhao, Wei, and Wen}]{Wang_2024}
Lei Wang, Chen Ma, Xueyang Feng, Zeyu Zhang, Hao Yang, Jingsen Zhang, Zhiyuan Chen, Jiakai Tang, Xu~Chen, Yankai Lin, Wayne~Xin Zhao, Zhewei Wei, and Jirong Wen. 2024{\natexlab{a}}.
\newblock \href {https://doi.org/10.1007/s11704-024-40231-1} {A survey on large language model based autonomous agents}.
\newblock \emph{Frontiers of Computer Science}, 18(6).

\bibitem[{Wang et~al.(2004)Wang, Bovik, Sheikh, and Simoncelli}]{1284395}
Zhou Wang, A.C. Bovik, H.R. Sheikh, and E.P. Simoncelli. 2004.
\newblock \href {https://doi.org/10.1109/TIP.2003.819861} {Image quality assessment: from error visibility to structural similarity}.
\newblock \emph{IEEE Transactions on Image Processing}, 13(4):600--612.

\bibitem[{Wang et~al.(2024{\natexlab{b}})Wang, Yu, Stengel-Eskin, Yoon, Cheng, Bertasius, and Bansal}]{wang2024videotreeadaptivetreebasedvideo}
Ziyang Wang, Shoubin Yu, Elias Stengel-Eskin, Jaehong Yoon, Feng Cheng, Gedas Bertasius, and Mohit Bansal. 2024{\natexlab{b}}.
\newblock \href {https://arxiv.org/abs/2405.19209} {Videotree: Adaptive tree-based video representation for llm reasoning on long videos}.
\newblock \emph{Preprint}, arXiv:2405.19209.

\bibitem[{Wu et~al.(2023)Wu, Bansal, Zhang, Wu, Li, Zhu, Jiang, Zhang, Zhang, Liu, Awadallah, White, Burger, and Wang}]{wu2023autogenenablingnextgenllm}
Qingyun Wu, Gagan Bansal, Jieyu Zhang, Yiran Wu, Beibin Li, Erkang Zhu, Li~Jiang, Xiaoyun Zhang, Shaokun Zhang, Jiale Liu, Ahmed~Hassan Awadallah, Ryen~W White, Doug Burger, and Chi Wang. 2023.
\newblock \href {https://arxiv.org/abs/2308.08155} {Autogen: Enabling next-gen llm applications via multi-agent conversation}.
\newblock \emph{Preprint}, arXiv:2308.08155.

\bibitem[{Yamamoto et~al.(2003)Yamamoto, Ogata, and Ariki}]{DBLP:conf/interspeech/YamamotoOA03}
Natsuo Yamamoto, Jun Ogata, and Yasuo Ariki. 2003.
\newblock \href {https://doi.org/10.21437/EUROSPEECH.2003-333} {Topic segmentation and retrieval system for lecture videos based on spontaneous speech recognition}.
\newblock In \emph{8th European Conference on Speech Communication and Technology, {EUROSPEECH} 2003 - {INTERSPEECH} 2003, Geneva, Switzerland, September 1-4, 2003}, pages 961--964. {ISCA}.

\bibitem[{Yang and Meinel(2014)}]{6750040}
Haojin Yang and Christoph Meinel. 2014.
\newblock \href {https://doi.org/10.1109/TLT.2014.2307305} {Content based lecture video retrieval using speech and video text information}.
\newblock \emph{IEEE Transactions on Learning Technologies}, 7(2):142--154.

\bibitem[{Yang et~al.(2011{\natexlab{a}})Yang, Sack, and Meinel}]{Yang2011LectureVI}
Haojin Yang, Harald Sack, and Christoph Meinel. 2011{\natexlab{a}}.
\newblock \href {https://api.semanticscholar.org/CorpusID:14263115} {Lecture video indexing and analysis using video ocr technology}.
\newblock \emph{2011 Seventh International Conference on Signal Image Technology \& Internet-Based Systems}, pages 54--61.

\bibitem[{Yang et~al.(2011{\natexlab{b}})Yang, Siebert, Lühne, Sack, and Meinel}]{yang2011lecture}
Haojin Yang, Maria Siebert, Patrick Lühne, Harald Sack, and Christoph Meinel. 2011{\natexlab{b}}.
\newblock \href {https://doi.org/10.1109/SITIS.2011.20} {Lecture video indexing and analysis using video ocr technology}.

\bibitem[{Yu et~al.(2023)Yu, Cho, Yadav, and Bansal}]{yu2023selfchainedimagelanguagemodelvideo}
Shoubin Yu, Jaemin Cho, Prateek Yadav, and Mohit Bansal. 2023.
\newblock \href {https://arxiv.org/abs/2305.06988} {Self-chained image-language model for video localization and question answering}.
\newblock \emph{Preprint}, arXiv:2305.06988.

\bibitem[{Zhang et~al.(2023)Zhang, Li, and Bing}]{zhang-etal-2023-video}
Hang Zhang, Xin Li, and Lidong Bing. 2023.
\newblock \href {https://doi.org/10.18653/v1/2023.emnlp-demo.49} {Video-{LL}a{MA}: An instruction-tuned audio-visual language model for video understanding}.
\newblock In \emph{Proceedings of the 2023 Conference on Empirical Methods in Natural Language Processing: System Demonstrations}, pages 543--553, Singapore. Association for Computational Linguistics.

\bibitem[{Zhao et~al.(2023)Zhao, Zhou, Li, Tang, Wang, Hou, Min, Zhang, Zhang, Dong et~al.}]{zhao2023survey}
Wayne~Xin Zhao, Kun Zhou, Junyi Li, Tianyi Tang, Xiaolei Wang, Yupeng Hou, Yingqian Min, Beichen Zhang, Junjie Zhang, Zican Dong, et~al. 2023.
\newblock A survey of large language models.
\newblock \emph{arXiv preprint arXiv:2303.18223}.

\end{thebibliography}
